%% file: arxiv.tex
\documentclass[11pt]{article}

\usepackage{times}
\usepackage[T1]{fontenc}
\usepackage{amsmath,amssymb,amsthm,mathtools}
\usepackage{booktabs,array,tabularx,microtype}
\usepackage{graphicx}
\usepackage{enumitem}
\usepackage{algorithm}
\usepackage{algpseudocode}
\usepackage{needspace}
\usepackage{xcolor}
\usepackage{natbib}
\usepackage[toc,page,header]{appendix}
\usepackage[colorlinks=true,linkcolor=blue,citecolor=blue,urlcolor=black,linktocpage=true]{hyperref}
\usepackage{url}
\usepackage{aliascnt}
\usepackage[capitalize,noabbrev,nameinlink]{cleveref}
\hypersetup{
  pdftitle={Reverse-KL Policy Distillation: Statistical Guarantees and Limits},
  pdfauthor={Haolin Liu, Chen-Yu Wei, Julian Zimmert},
  pdfsubject={Reverse-KL policy distillation}
}
\allowdisplaybreaks[2]
\newtheorem{theorem}{Theorem}[section]
\newaliascnt{lemma}{theorem}
\newtheorem{lemma}[lemma]{Lemma}
\aliascntresetthe{lemma}
\newaliascnt{proposition}{theorem}
\newtheorem{proposition}[proposition]{Proposition}
\aliascntresetthe{proposition}
\newaliascnt{corollary}{theorem}
\newtheorem{corollary}[corollary]{Corollary}
\aliascntresetthe{corollary}
\theoremstyle{definition}
\newtheorem{assumption}{Assumption}[section]
\crefname{assumption}{Assumption}{Assumptions}
\Crefname{assumption}{Assumption}{Assumptions}
\newaliascnt{definition}{theorem}

\aliascntresetthe{definition}
\theoremstyle{remark}
\newaliascnt{remark}{theorem}

\aliascntresetthe{remark}
\newcommand{\KL}{\operatorname{KL}}
\newcommand{\E}{\mathbb E}
\newcommand{\Prob}{\mathbb P}
\newcommand{\1}{\mathbf 1}
\newcommand{\Hell}{D_{\mathrm H}}
\newcommand{\Ber}{\operatorname{Ber}}
\newcommand{\Rcal}{\mathcal R}
\newcommand{\Capx}{C_{\rm app}}

\title{Reverse-KL Policy Distillation: \\Statistical Guarantees and Limits}
\author{%
  Haolin Liu\\
  \scalebox{0.9}{University of Virginia}\\
  \scalebox{0.9}{\texttt{srs8rh@virginia.edu}}
  \and
  Chen-Yu Wei\\
  \scalebox{0.9}{University of Virginia}\\
  \scalebox{0.9}{\texttt{chenyu.wei@virginia.edu}}
  \and
  Julian Zimmert\\
  \scalebox{0.9}{Google Research /}\\
  \scalebox{0.9}{National Taiwan University}\\
  \scalebox{0.9}{\texttt{zimmert@google.com}}
}
\date{}

\begin{document}
\addtocontents{toc}{\protect\setcounter{tocdepth}{-10}}
\maketitle

\input{sections/abstract}
\input{sections/introduction}
\input{sections/problem_formulation}

\input{sections/online_learning}
\input{sections/offline_learning}
\input{sections/conclusion}

\bibliographystyle{plainnat}
\bibliography{ref}

\clearpage
\appendix
\appendixpage
\addtocontents{toc}{\protect\setcounter{tocdepth}{3}}
\setcounter{tocdepth}{3}
\makeatletter
\@starttoc{toc}
\makeatother

\input{appendices/related_work}
\input{appendix_online}
\input{appendix_offline}
\input{appendix_auxiliary}

\addtocontents{toc}{\protect\setcounter{tocdepth}{-10}}

\input{appendices/reserved_draft}
\end{document}

%% file: sections/abstract.tex
\begin{abstract}
  Reverse Kullback--Leibler (KL) distillation has shown empirical promise for training
  language models, but its statistical guarantees and limits remain poorly understood. 
  In particular, the effects of two design choices remain unclear: Does access to the teacher's full action distribution provide an advantage over observing only the probability it assigns to the sampled action?  Must training trajectories come from the student (online distillation), or can teacher-generated trajectories (offline distillation) suffice?  Our work takes a first step toward understanding the theory underlying these questions.


  We study the sample complexity of online reverse-KL distillation under two feedback models: the teacher reveals either its full action distribution or only the probability of the \emph{student}-generated action. We establish minimax bounds for the former and show that substantially more data may be required for the latter. 
In contrast, offline reverse-KL distillation can attain the same minimax rate whether the teacher reveals its full distribution or only its probability of the \emph{teacher}-generated action.
The key distinction is that teacher-generated actions themselves reveal the teacher's preferences, while queries at student-selected actions can leave those preferences unobserved. This highlights an additional statistical challenge for online distillation compared to offline distillation. 

These statistical guarantees do not follow simply from optimizing standard distillation objectives: our lower bounds show that common online objectives can fail when candidates are evaluated only on past student-generated trajectories.
We develop \emph{trajectory-loss} and \emph{importance-ratio} truncation techniques together with policy-space learning to achieve optimal online distillation guarantees. 

\end{abstract}

%% file: sections/introduction.tex
\section{Introduction}\label{sec:intro}

Policy distillation compresses a teacher into a student that may lack the capacity
to reproduce the teacher's behavior in full.
The choice of divergence therefore matters.
Forward Kullback--Leibler (KL) divergence (i.e., $\KL$(teacher$\|$student)) encourages the student
to cover the teacher's full range of responses.
Reverse KL (i.e., $\KL$(student$\|$teacher)) instead allows a limited-capacity student to concentrate
on teacher behaviors it can represent, while discouraging responses
that are unlikely under the teacher.
Reverse KL has shown empirical promise in on-policy distillation for large language models (LLMs), where the
student generates responses and learns from the teacher's action probability
 at the observed state \citep{gu2024,agarwal2024,lu2025}.

Although the empirical results are promising, the optimal statistical guarantees
for reverse-KL learning remain unclear.
The statistical question depends first on what information the teacher provides.
Some distillation methods only use the teacher's probability on
student-generated actions \citep{lu2025}, whereas recent studies report benefits
from access to the teacher's full action distribution
\citep{li2026policy,yu2026mismatch,fu2026}. Whether this richer form of teacher
feedback improves statistical efficiency, and by how much, remains open. Moreover, reverse KL is evaluated under the student's distribution, so training typically
uses student-generated responses.
It is therefore unclear whether learning from offline teacher-generated data can match or even improve on the
statistical guarantees attainable from online student rollouts.

To answer these questions, we study sequence-level reverse-KL learning with a student policy class that may not contain the teacher, allowing for
\emph{misspecification}.
Our framework distinguishes two aspects of the learning process: who generates
the training trajectories and what information the teacher provides along them.
The trajectories come either from student policies chosen adaptively by the
learner in the \emph{online} setting, or from the teacher in the \emph{offline}
setting.
With either data source, the teacher may reveal its entire action distribution
at each visited state (\emph{full feedback}) or only the probability of the
observed action (\emph{action feedback}).
Across these four settings, we study how closely a learner can
match the best student in its policy class for a given number
of training trajectories.

We give a rather complete picture for this problem.
With full feedback, we establish the minimax-optimal guarantees for reverse-KL
learning under a relative second-moment assumption. For the online setting, 
we first show that common local distillation losses can
be suboptimal.
These objectives, motivated by DAgger and on-policy distillation
\citep{ross2011,lu2025}, evaluate candidate policy on states generated by the
rollout policies, including state that a candidate itself would rarely
visit, which can distort policy rankings. 
Our algorithm (\cref{alg:online}) corrects this
mismatch by reweighting the observed prefix to account for each policy's own
generation behavior, attaining the optimal guarantee
(\cref{sec:online}).
For the offline setting, we 
develop a new algorithm (\cref{eq:offline-full-estimator}) that attains the
same optimal guarantee from teacher-generated trajectories.

With action feedback only, online learning can require substantially more samples
than with full feedback (\cref{sec:online-action}).
Yet under a trajectory likelihood-ratio bound, offline learning with action feedback can still obtain a comparable guarantee to full feedback learning (\cref{sec:online-action,sec:offline}).
These results show that richer feedback can improve online sample efficiency,
while surprisingly optimal reverse-KL learning need not require online interaction.

Formally, let $\pi^\star$ be the teacher policy, $\Pi$ be the
student policy class, and $P^\pi$ the sequence distribution generated by $\pi$.
Our guarantees take the form
\[
  \KL(P^{\widehat\pi}\|P^{\pi^\star})
  \le \Capx\min_{\pi\in\Pi}\KL(P^\pi\|P^{\pi^\star})
  +C_{\rm est}\frac{\log(|\Pi|/\delta)}{T}
\]
with probability at least $1-\delta$, using $T$ complete episodes and returning
$\widehat\pi\in\Pi$.
Here $\Capx$ (approximation factor) measures the approximation cost of competing with the best available student,
and $C_{\rm est}$ (estimation factor)  controls the estimation error.
At the minimax scale of $C_{\rm est}$, we characterize the smallest attainable
$\Capx$.
\cref{tab:feedback} summarizes the guarantees and separations.

\begin{table}[H]
  \centering
  \caption{Reverse-KL learning by data source and teacher feedback.
    Approximation coefficients are compared at the optimal estimation scale.}
  \label{tab:feedback}
  \small
  \setlength{\tabcolsep}{4pt}
  \renewcommand{\arraystretch}{1.25}
  \begin{tabularx}{\linewidth}{@{}l l >{\centering\arraybackslash}X c c@{}}
    \toprule
    Data source & Feedback & Assumption & $\Capx$ & $C_{\rm est}$ \\
     \midrule
    Online / Offline & Full
    & \shortstack{Relative second moment\\[2pt]
      $\E_{P^\pi}[K_\pi^2]\le B\,\E_{P^\pi}[K_\pi]$}
    & $\Theta(B)$ & $\Theta(B)$ \\[5pt]
    Offline & Action
    & $P^\pi(\tau)\le e^{B_\infty}P^{\pi^\star}(\tau)$
    & $\Theta(B_\infty)$ & $\Theta(B_\infty)$ \\[5pt]
    Online & Action
    & $P^\pi(\tau)\le e^{B_\infty}P^{\pi^\star}(\tau)$
    & --- & $\Omega\!\left(\frac{B_\infty m}{\log(em)}\right)$ \\
    \bottomrule
  \end{tabularx}
  \par\smallskip
  \begin{minipage}{\linewidth}
    \small
    Here $K_\pi(\tau)=\sum_{h=1}^H\KL(\pi_h(\cdot\mid s_h)\|\pi_h^\star(\cdot\mid s_h))$ for every feasible trajectory $\tau = (s_{1:H}, a_{1:H})$,
    and $m=\min\{|\mathcal A|,|\Pi|,e^{B_\infty}\}$. The assumptions hold for every $\pi$ and feasible trajectory $\tau$. The trajectory likelihood-ratio condition is stronger and implies the relative second moment condition with
    $B=4(B_\infty+1)$ (\cref{lem:second-moment-sufficient}).
  \end{minipage}
\end{table}

\paragraph{Related work.} Our closest theoretical predecessors  \citet{rohatgi2025} and \citet{xu2026} also study imitation learning under misspecification. 
\citet{rohatgi2025} obtain constant approximation and estimation
coefficients for learning trajectory Hellinger distance with computationally inefficient algorithms, and show that standard next-token prediction approaches (e.g. cross-entropy loss) must suffer an $\Omega(H)$ approximation factor where $H$ is the learning horizon. \citet{xu2026} establish a constant approximation factor for learning forward KL divergence under standard cross-entropy loss, but the estimation error scales with a boundedness factor. Both papers only study offline learning from teacher trajectories with action
feedback. We study reverse-KL learning in both online and offline settings
and characterize the statistical advantage of full feedback
over action feedback.
\cref{app:related-work} provides further discussion of related work.

%% file: sections/problem_formulation.tex
\section{Problem formulation}\label{sec:setup}

\paragraph{Notation.}
For a positive integer $n$, write $[n]=\{1,\ldots,n\}$.
All logarithms are natural unless a base is displayed.
We use $x\wedge y=\min\{x,y\}$, $x_+=\max\{x,0\}$, and $\1\{E\}$ for the indicator of an
event $E$.
The notation $O$, $\Omega$, and $\Theta$ denotes upper, lower, and two-sided bounds up
to positive numerical factors, and $\lesssim$ suppresses such an upper-bound factor. We write $x\asymp y$ when both $x\lesssim y$ and $y\lesssim x$. For distributions $p,q$,
write $\KL(p\|q)=\sum_z p(z)\log(p(z)/q(z))$, where terms with $p(z)=0$
equal zero and $\KL(p\|q)=+\infty$ if $p(z)>0$ and $q(z)=0$ for some $z$.

\paragraph{Setting.} We consider a reward-free Markov decision process with horizon
$H\ge1$, discrete state space $\mathcal S$, action set $\mathcal A$, initial state 
distribution $\mu$, and transition kernel $\mathsf P_h(\cdot\mid s,a)$.
A Markov policy $\pi=(\pi_h)_{h=1}^H$ consists of maps
$\pi_h:\mathcal S\to\Delta(\mathcal A)$, where
$\Delta(\mathcal A)=\{p\in[0,1]^{|\mathcal A|}:\sum_{a\in\mathcal A}p(a)=1\}$
is the probability simplex over actions. We write $\pi_h(a\mid s)$ for the
probability assigned to action $a$ at state $s$. Let $s_h\in\mathcal S$ and
$a_h\in\mathcal A$ denote the state and action at step $h$. Under $\pi$, an
episode generates a trajectory
$\tau=(s_1,a_1,\ldots,s_H,a_H)\in\mathcal T=(\mathcal S\times\mathcal A)^H$
by sampling $s_1\sim\mu$, $a_h\sim\pi_h(\cdot\mid s_h)$, and, for $h<H$,
$s_{h+1}\sim\mathsf P_h(\cdot\mid s_h,a_h)$. We write $P^\pi$ for the 
trajectory distribution under policy $\pi$, so
\(
  P^\pi(\tau)
  =\mu(s_1)\prod_{h=1}^H\pi_h(a_h\mid s_h)
    \prod_{h=1}^{H-1}\mathsf P_h(s_{h+1}\mid s_h,a_h).
\)
Let $\xi_h=(s_1,a_1,\ldots,s_{h-1},a_{h-1},s_h)$ denote the prefix
before action $a_h$. Its probability under $\pi$ is
\(
  P^\pi(\xi_h)
  =\mu(s_1)\prod_{j=1}^{h-1}
    \pi_j(a_j\mid s_j)\mathsf P_j(s_{j+1}\mid s_j,a_j).
\)
The empty product equals one when $h=1$.

The learner is given a finite policy class
$\Pi$ and seeks to imitate a fixed teacher
$\pi^\star$. 
Define the local reverse KL between policy $\pi$ and $\pi^\star$ at state $s$ and step $h$ by
$k_h(s;\pi)=\KL(\pi_h(\cdot\mid s)\|\pi_h^\star(\cdot\mid s))$.
Our target is the \emph{trajectory reverse-KL risk}, whose chain-rule decomposition is
\(
  L(\pi):=\KL(P^\pi\|P^{\pi^\star})
  =\E_{P^\pi}\big[\sum_{h=1}^H k_h(s_h;\pi)\big].
\)
Let $L^\star=\min_{\pi\in\Pi}L(\pi)$.
The setting is \emph{realizable} when $\pi^\star\in\Pi$, giving $L^\star=0$, and
\emph{agnostic} when the teacher need not belong to the class.

\paragraph{Observation models.} With a budget of $T$ complete episodes, we consider four observation models (see \cref{tab:feedback}) that differ in their \emph{trajectory source} and \emph{teacher feedback}:  
\begin{itemize}[
  leftmargin=1.5em,
  itemsep=0.2em,
  topsep=0em,
  partopsep=0pt,
  parsep=0pt
]
\item Trajectory source: 
In the \emph{online} setting, at each round $t\in[T]$, the learner selects
$\pi_t\in\Pi$ based on past observations and executes it for a complete episode,
obtaining $\tau_t\sim P^{\pi_t}$. In the \emph{offline} setting, the learner
receives independent teacher trajectories
$\tau_t^\star\sim P^{\pi^\star}$ for $t\in[T]$.
\item Teacher feedback: 
At each visited state $s_h$, the learner receives one of two forms of teacher
feedback. \emph{Full feedback} reveals the teacher's entire action distribution
$\pi_h^\star(\cdot\mid s_h)$, whereas \emph{action feedback} reveals only the
probability $\pi_h^\star(a_h\mid s_h)$ assigned to the observed action $a_h$.
Either feedback model can accompany online or offline trajectories, and no
feedback is available at unvisited states.
\end{itemize}

\paragraph{Boundedness assumptions.}
Our full-feedback guarantees hold under a relative second-moment
condition. For action feedback, we use a stronger
trajectory likelihood-ratio bound.

\begin{assumption}[Relative second-moment bound]
\label{ass:relative-second-moment}
There exists $B\ge1$ such that, for every $\pi\in\Pi$,
\begin{equation*}
\E_{\tau\sim P^\pi}\!\left[
  \left(\sum_{h=1}^H
    \KL\!\left(\pi_h(\cdot\mid s_h)\,\middle\|\,
      \pi_h^\star(\cdot\mid s_h)\right)\right)^2
\right]
\le B\,\KL(P^\pi\|P^{\pi^\star}) < \infty.
\end{equation*}
\end{assumption}
By the KL chain rule, the average cumulative local KL under
$P^\pi$ equals the trajectory divergence
$\KL(P^\pi\|P^{\pi^\star})$.
\cref{ass:relative-second-moment} thus imposes a
\emph{Bernstein-type moment condition}, the second moment is
controlled linearly by the mean under each candidate
policy's own trajectory distribution.
Intuitively, trajectories with cumulative local KL much larger
than $B$ can contribute only a small fraction of the average
reverse KL.

This condition is weaker than several uniform boundedness
assumptions. For example, a uniform local KL bound $k_h(s;\pi)\le G$ for all $h,s,\pi$  implies \cref{ass:relative-second-moment} with $B=\max\{1,HG\}$.
The trajectory likelihood-ratio bound in
\cref{ass:ratio-cap} below also implies \cref{ass:relative-second-moment}, with
$B=4(B_\infty+1)$.
\cref{app:second-moment} provides the precise implications
and proofs.

Let
$\mathcal T_{\rm env}$ denote the set of environment-feasible trajectories
$\tau=(s_{1:H},a_{1:H})$ satisfying
$\mu(s_1)\prod_{h=1}^{H-1}\mathsf P_h(s_{h+1}\mid s_h,a_h)>0$. \cref{ass:ratio-cap} gives trajectory likelihood-ratio bound.
\begin{assumption}[Trajectory likelihood-ratio bound]\label{ass:ratio-cap}
  There is $B_\infty>1$ such that
  $P^{\pi}(\tau)\le e^{B_\infty}P^{\pi^\star}(\tau)$ for every
  $\pi\in\Pi$ and $\tau\in\mathcal T_{\rm env}$.
\end{assumption}
The local bound
$\pi_h(a\mid s)\le e^G\pi_h^\star(a\mid s)$ implies trajectory-level bound
$P^\pi(\tau)\le e^{HG}P^{\pi^\star}(\tau)$.
Thus, $B$ and $B_\infty$ may themselves grow with $H$,
so the absence of an explicit horizon factor in our guarantees
does not imply horizon independence.

Like our \cref{ass:ratio-cap}, Assumption~1 of
\citet{xu2026} also bounds likelihood ratios, but in the
opposite direction $P^{\pi^\star}(\tau)/P^\pi(\tau)$ rather than our
$P^\pi(\tau)/P^{\pi^\star}(\tau)$. Their assumption requires every student to cover the
teacher for a forward-KL objective, whereas our 
\cref{ass:ratio-cap} requires the teacher to cover every student for a reverse KL objective. Teacher coverage is more natural in many cases because a limited-capacity student
may concentrate on teacher-supported behaviors without representing the teacher’s entire range of
behaviors.

\paragraph{Approximation error and estimation error.} Following \citet{rohatgi2025} and \citet{xu2026}, we seek a learner that returns
$\widehat\pi\in\Pi$ such that, with probability at least $1-\delta$,
\[
  L(\widehat\pi)\le
  \underbrace{\Capx L^\star}_{\text{approximation error}}
  +
  \underbrace{C_{\rm est}\frac{\log(|\Pi|/\delta)}{T}}_{\text{estimation error}}.
\]
The coefficients may depend on $B,B_\infty,H,|\Pi|,|\mathcal A|$ and the observation
model.
Here $L^\star$ is the smallest risk achievable by any student in $\Pi$.
The coefficient $\Capx$ measures how closely the learner competes with this
best-in-class risk. The coefficient $C_{\rm est}$ scales the estimation term
$\log(|\Pi|/\delta)/T$, determining how many episodes are needed for accurate
learning. We fix $C_{\rm est}$ at its minimax-optimal order and characterize the
smallest attainable $\Capx$.

%% file: sections/online_learning.tex
\section{Online learning from student trajectories}\label{sec:online}

In this section, we study online reverse-KL learning, where student policies
generate trajectories and the teacher reveals its full action distribution at
each visited state. We first show that common
distillation losses directly evaluate any candidate policy on trajectories generated by past behavior policies without any distribution mismatch correction. Such objectives can penalize a candidate policy for errors at states it would rarely visit itself, leading to suboptimal guarantees.  To tackle this limitation, we then introduce a new algorithm that corrects this mismatch and attains the optimal guarantee.

\subsection{Why common objectives can fail}
A natural approach fits each new student to the teacher's predictions on
states visited during training, as in DAgger \citep{ross2011} and standard on-policy distillation \citep{agarwal2024}.
At round $t$, the current student $\pi_t$ generates a batch of  trajectories with full teacher feedback. Let $\mathcal D_t$ contain all
trajectories observed till round $t$. The learner chooses the next policy by
\begin{equation}\label{eq:empirical-refitting}
  \pi_{t+1}\in\operatorname*{arg\,min}_{\pi\in\Pi}Q_t(\pi;\mathcal D_t).
\end{equation}
where $Q_t$ is the objective at round $t$, which may use any nonempty subset of $\mathcal D_t$. For example, reverse-KL distillation minimizes 
$Q_t(\pi;\mathcal D_t) = \sum_{h=1}^H k_h(s_h;\pi)$ over the selected trajectories $(s_{1:H}, a_{1:H}) \in \mathcal{D}_t$. Such local
reverse-KL losses appear in generalized knowledge distillation and recent
on-policy distillation studies \citep{agarwal2024,li2026}.
Other choices include forward KL, mixtures of the two KL directions, and
clipped KL losses \citep{lin2020,wu2025,zhao2026}.
\cref{app:training-failure} specifies the general $Q_t$ objective we consider, including all objectives discussed above.  \cref{prop:training-failure} shows that this class of algorithms can suffer high suboptimality gap.

\begin{proposition}\label{prop:training-failure}
For every sufficiently large $B$, there is an $H=2$, $|\Pi|=2$ instance
satisfying \cref{ass:relative-second-moment,ass:ratio-cap} with $B_\infty=2B$ and
$L^\star\le e^{-\Omega(B)}$. For algorithm in \cref{eq:empirical-refitting} with $Q_t$ satisfies the constraints in \cref{app:training-failure}, under uniform random initialization, after $T=\Theta(B^3)$ episodes,
the returned policy has risk $\Omega(B)$ with
constant probability.
\end{proposition}
The construction has a high-risk policy $\pi_{\rm bad}$ that frequently
visits a state where the near-optimal policy $\pi_{\rm good}$ has a large  loss. Although $\pi_{\rm good}$ rarely reaches this state itself, the loss on trajectories from $\pi_{\rm bad}$ makes $\pi_{\rm good}$ look worse. Thus, with constant probability, random initialization selects $\pi_{\rm bad}$
and every subsequent policy optimization retains it. To address this problem, our algorithm reduces the influence of states that the evaluated
policy would rarely reach.

\subsection{Correcting visitation mismatch}

Suppose $\pi_t$ generates a trajectory $(s_{1:H}, a_{1:H})$ at round $t \in [T]$. Full teacher feedback reveals the
local loss $k_h(s_h;\pi)$ for every candidate $\pi$ at each visited state, but
the visited prefixes are sampled according to $\pi_t$ rather than $\pi$.
Writing $\xi_h=(s_1,a_1,\ldots,s_{h-1},a_{h-1},s_h)$ for the prefix before
action $a_h$, standard importance weighting multiplies a loss at $\xi_h$ by
$P^\pi(\xi_h)/P^{\pi_t}(\xi_h)$. Under rollouts from $\pi_t$, this changes its
expected coefficient from $P^{\pi_t}(\xi_h)$ to $P^\pi(\xi_h)$. However, the importance ratio can be arbitrarily large
when a prefix is rare under $\pi_t$, making the resulting estimate highly sensitive
to rare observations.

We therefore clip the weight at one, giving
$\min\{1,P^\pi(\xi_h)/P^{\pi_t}(\xi_h)\}$. Any loss attached to $\xi_h$ then enters
the expectation with coefficient
$P^{\pi_t}(\xi_h)\min\{1,P^\pi(\xi_h)/P^{\pi_t}(\xi_h)\}
=\min\{P^\pi(\xi_h),P^{\pi_t}(\xi_h)\}$.
Thus, a prefix $\xi_h$ that $\pi$ visits less often than $\pi_t$ is downweighted to
$P^\pi(\xi_h)$, whereas a prefix that $\pi$ visits more often is not amplified
beyond $P^{\pi_t}(\xi_h)$.
The ratio can be computed as $P^\pi(\xi_h)/P^{\pi_t}(\xi_h)
=\prod_{u<h}\pi_u(a_u | s_u)/\pi_{t,u}(a_u | s_u)$ because the
transition probabilities cancel. 

Although each importance ratio is clipped, the local KL losses along a
trajectory can still sum to a large value. We therefore consider the clipped cumulative local loss $\min\!\big\{1,\sum_{h=1}^H k_h(s_h;\pi)\big\}$. To combine this
trajectory-level clipping with the step-dependent prefix weights, we decompose
the clipped total loss into increments. Specifically, let $e_h(\xi_h;\pi)$ be the
increase caused by adding the local loss at step $h$,
\begin{equation}\label{eq:increments}
  e_h(\xi_h;\pi)=\min\!\left\{1,\sum_{u=1}^{h}k_u(s_u;\pi)\right\}
  -\min\!\left\{1,\sum_{u=1}^{h-1}k_u(s_u;\pi)\right\}.
\end{equation}
The increments $e_h(\xi_h;\pi)$ are nonnegative and $\sum_{h=1}^H e_h(\xi_h;\pi)
  =\min\!\big\{1,\sum_{h=1}^H k_h(s_h;\pi)\big\}\le1$. 

The two clipping operations reduce the influence of rare, high-loss observations,
but they also discard information about the original reverse-KL risk. To
quantify what remains, let
$\overline L(\pi):=\E_{P^\pi}\big[\sum_{h=1}^H e_h(\xi_h;\pi)\big]$ denote policy $\pi$'s expected
clipped loss.

First, under \cref{ass:relative-second-moment}, clipping incurs a factor $O(B)$,
\begin{equation}\label{eq:visible}
  \overline L(\pi)\le L(\pi)\lesssim B\,\overline L(\pi).
\end{equation}
Thus, controlling the clipped loss $\overline L(\pi)$ suffices to control sequence risk $L(\pi)$, at a
cost of $B$ (\cref{lem:second-moment-clipping}). 

Second, the rollout policy $\pi_t$ may rarely visit prefixes that are important to $\pi$. Since the importance weight is clipped, trajectories from $\pi_t$
cannot fully recover $\pi$'s loss on prefixes $\xi_h$ with
$P^{\pi_t}(\xi_h)<P^\pi(\xi_h)$. The clipped importance sampling ratios retain each increment $e_h(\xi_h;\pi)$
with probability mass
$\min\{P^\pi(\xi_h),P^{\pi_t}(\xi_h)\}$
rather than its full mass $P^\pi(\xi_h)$ in $\overline L(\pi)$.
More generally, for any $\pi,\pi'\in\Pi$, we define
\begin{equation}\label{eq:truncated-clipped-objective}
  \scalebox{0.97}{$\displaystyle
    \operatorname{Trun}_{\pi'}\left[\overline L(\pi)\right] :=\E_{P^{\pi'}}\!\left[\sum_{h=1}^H
    \min\!\left\{1,\frac{P^\pi(\xi_h)}{P^{\pi'}(\xi_h)}\right\}e_h(\xi_h;\pi)\right] =\sum_{h,\xi_h}\min\{P^\pi(\xi_h),P^{\pi'}(\xi_h)\}\,e_h(\xi_h;\pi).$
  }
\end{equation}
Here, $\pi'$ in the subscript is the rollout policy, while $\pi$ inside
$\overline L(\pi)$ is the policy being evaluated. In episode $t$, we take
$\pi'=\pi_t$.
Since $\operatorname{Trun}_{\pi_t}[\overline L(\pi)]$ retains only the
prefix mass shared by $\pi$ and $\pi_t$, it may underestimate
$\overline L(\pi)$. Indeed,
$0\le\operatorname{Trun}_{\pi_t}[\overline L(\pi)]\le\overline L(\pi)$.
The key fact is that when $\overline L(\pi_t)$ is small, the truncated
quantity still controls $\overline L(\pi)$. \cref{lem:visibility-main}
gives 
\begin{equation}\label{eq:recover-clipped-objective}
  \overline L(\pi)\le64\left(
  \operatorname{Trun}_{\pi_t}\left[\overline L(\pi)\right]
  +\overline L(\pi_t)\right).
\end{equation}
\cref{eq:recover-clipped-objective} shows why the quality of
the rollout policy matters. When $\overline L(\pi_t)$ is small, a
candidate $\pi$ with substantially larger $\overline L(\pi)$ must
also have a large 
$\operatorname{Trun}_{\pi_t}\left[\overline L(\pi)\right]$.
Thus, even after truncation, losses observed on rollouts from $\pi_t$
provide evidence against $\pi$.
Since the learner does not know in advance how to choose a rollout policy
$\pi_t$ with small $\overline L(\pi_t)$, the observed losses must guide both candidate evaluation
and the choice of future rollout policies.

\subsection{Online learning for reverse KL minimization}
\label{sec:onlin alg}

To let the observed losses guide future rollouts, our algorithm assigns
each candidate $\pi$ a weight $w_t(\pi)$ and samples $\pi_t\sim w_t$ in
episode $t$.
After executing $\pi_t$, the learner observes the trajectory $\tau_t = (s_{t,1:H}, a_{t,1:H})$ and full teacher
feedback $\pi^\star(\cdot|s_{t,h})$ for every $h \in [H]$. These observations are used to compare every $\pi\in\Pi$ with $\pi_t$.
At the observed prefix $\xi_{t,h} = (s_{t,1}, a_{t,1}, \cdots, s_{t,h})$, the loss $\ell_t(\pi)$ subtracts
$e_h(\xi_{t,h};\pi_t)$ from $e_h(\xi_{t,h};\pi)$ and applies the same
clipped prefix importance sampling weight to both terms, yielding
\begin{equation}\label{eq:online-loss}
  \ell_t(\pi)=\sum_{h=1}^H
  \min\!\left\{1,\frac{P^\pi(\xi_{t,h})}{P^{\pi_t}(\xi_{t,h})}\right\}
  \bigl(e_h(\xi_{t,h};\pi)-e_h(\xi_{t,h};\pi_t)\bigr).
\end{equation}
Let $\E_t$ denote expectation conditioned on the history before round $t$,
averaging over both $\pi_t$ and its trajectory.  We have 
\begin{align}
\E_t\left[\ell_t(\pi)\right] = \E_{\pi_t \sim w_t}\left[\operatorname{Trun}_{\pi_t}\left[\overline L(\pi)\right]
-\operatorname{Trun}_{\pi}\left[\overline L(\pi_t)\right]\right]. 
\label{eq:exp-trun}  
\end{align}
Thus in expectation, $\ell_t(\pi)$ compares the losses of $\pi$ and $\pi_t$ under  prefix probability mass $\min\{P^\pi(\xi_{t,h}),\allowbreak P^{\pi_t}(\xi_{t,h})\}$. From clipping, each of the two weighted sums in \cref{eq:online-loss} lies in
$[0,1]$, so $\ell_t(\pi)\in[-1,1]$.

Our algorithm is \cref{alg:online}. The exponential-weights update turns
$\ell_t(\pi)$ into a change in sampling probability.
Since $\ell_t(\pi_t)=0$, a positive $\ell_t(\pi)$ decreases the weight
of $\pi$ relative to $\pi_t$, while a negative value increases the weight of $\pi$.
Candidates with smaller accumulated losses
$\sum_{u=1}^t\ell_u(\pi)$ receive larger $w_{t+1}(\pi)$ and are
therefore more likely to generate subsequent trajectories.
The algorithm outputs the policy with the largest final weight $w_{T+1}$.

\begin{algorithm}[H]
  \caption{Online learning by bounded prefix comparisons}\label{alg:online}
  \small
  \begin{algorithmic}[1]
    \Require Policy class $\Pi$, episode budget $T$.
    \State Set $w_1(\pi)=1/|\Pi|$ for all $\pi\in\Pi$ and $\eta=1/256$.
    \For{$t=1,\ldots,T$}
      \State Sample $\pi_t\sim w_t$ and execute it for the entire episode.
      \State Observe the trajectory $\tau_t = (s_{t,1:H}, a_{t, 1:H})$ and the teacher distribution $\pi^\star(\cdot|s_{t,h})$ for every $h \in [H]$.
      \State For every $\pi\in\Pi$, compute the  loss $\ell_t(\pi)$ using \cref{eq:online-loss}.
      \State $\displaystyle
        w_{t+1}(\pi)=\frac{w_t(\pi)\exp(-\eta\ell_t(\pi))}{\sum_{\pi' \in \Pi}
        w_t(\pi')\exp(-\eta\ell_t(\pi'))}$ for all $\pi\in\Pi$.
    \EndFor
    \State Return $\widehat\pi\in\operatorname*{arg\,max}_{\pi\in\Pi}w_{T+1}(\pi)$,
      breaking ties by the smallest candidate index.
  \end{algorithmic}
\end{algorithm}
The following theorem shows that \cref{alg:online} achieves the
optimal estimation scale while competing with $L^\star$.

\begin{theorem}[Online full-feedback guarantee]\label{thm:online}
  Under \cref{ass:relative-second-moment}, for every $\delta\in(0,1)$,
  \cref{alg:online} returns a candidate satisfying,
  with probability at least $1-\delta$,
  \begin{equation}\label{eq:online-bound}
    L(\widehat\pi)\le
      O\!\left(B\left[L^\star+\frac{\log(|\Pi|/\delta)}{T}\right]\right).
  \end{equation}
  The $O(\cdot)$ notation hides only universal numerical constants.
\end{theorem}

Under \cref{ass:ratio-cap}, \cref{lem:second-moment-sufficient}
gives \cref{ass:relative-second-moment} with $B=4(B_\infty+1)$,
so \cref{eq:online-bound} holds with $B_\infty$ replacing $B$. \cref{thm:online-optimality} establishes the tightness of both coefficients
in \cref{eq:online-bound}.

\begin{theorem}[Optimality of full-feedback learning]\label{thm:online-optimality}
For both online and offline learning with full feedback, the minimax
estimation coefficient is $C_{\mathrm{est}}=\Theta(B)$ under
\cref{ass:relative-second-moment}  in the realizable setting.
Any learner achieving $C_{\mathrm{est}}=O(B)$ on all instances
must incur $C_{\mathrm{app}}=\Omega(B)$ in the worst case.
Under \cref{ass:ratio-cap}, the same conclusions hold with
$B_\infty$ replacing $B$.
\end{theorem}

Together, \cref{thm:online,thm:online-optimality}
show that \cref{alg:online} attains the optimal estimation
rate and the smallest possible order of approximation coefficient at
that rate. The estimation and approximation lower bounds are proved in
\cref{app:estimation,app:proper}, respectively. Below, we give a proof sketch for the upper bound. 

\begin{proof}[Proof sketch of \cref{thm:online}]
 
By the standard second-order
exponential-weights analysis 
\citep[Eq.~(27.2)--(27.3)]{lattimore2020bandit}, 
for a fixed minimizer $\pi^\dagger\in\arg\min_{\pi\in\Pi}L(\pi)$,
$\ell_t(\pi)\in[-1,1]$, and $\eta\le1$, we have
\begin{align}
- \underbrace{\sum_{t=1}^T\ell_t(\pi^\dagger)}_{\textbf{benchmark loss}}
\le \frac{\log|\Pi|}{\eta}
- \underbrace{\sum_{t=1}^T\sum_{\pi\in\Pi}w_t(\pi)\ell_t(\pi)}_{\textbf{weighted loss}}
+ \underbrace{\eta\sum_{t=1}^T\sum_{\pi\in\Pi}
 w_t(\pi)\ell_t(\pi)^2}_{\textbf{quadratic term}}.
\label{eq:exp bound}
\end{align}
We further bound the conditional expectation of the three terms above. For the \textbf{benchmark loss},  
\begin{align}
    \E_t\!\left[\ell_t(\pi^\dagger)\right]
&=\E_{\pi_t \sim w_t}\left[
\operatorname{Trun}_{\pi_t}\left[\overline L(\pi^\dagger)\right]
-
\operatorname{Trun}_{\pi^\dagger}\left[\overline L(\pi_t)\right]\right]  \tag{\cref{eq:exp-trun}} 
\\&\le L^\star+\overline L(\pi^\dagger)
-\frac1{64}\E_{\pi_t\sim w_t}\left[\overline L(\pi_t)\right]
\tag{$\operatorname{Trun}_{\pi_t}\left[\overline L(\pi^\dagger)\right]
\le\overline L(\pi^\dagger)\le L^\star$ and \cref{eq:recover-clipped-objective}}\\
&\le 2L^\star-\frac1{64}
\sum_{\pi\in\Pi}w_t(\pi)\overline L(\pi).
\tag{$\overline L(\pi^\dagger)\le L^\star$}
\end{align}
The \textbf{weighted loss} has zero conditional expectation:   
\begin{align}
  \E_t\!\left[\sum_{\pi\in\Pi}w_t(\pi)\ell_t(\pi)\right] =\sum_{\pi,\pi_t\in\Pi}w_t(\pi)w_t(\pi_t)
\Bigl(
\operatorname{Trun}_{\pi_t}\left[\overline L(\pi)\right]
-\operatorname{Trun}_{\pi}\left[\overline L(\pi_t)\right]
\Bigr) =0.
\label{eq:weight}
\end{align}
For the \textbf{quadratic term}, since $e_h(\xi_{t,h};\pi) \in [0,1]$, we have 
\begin{align*}
   \E_t\!\left[\ell_t(\pi)^2\right]
&\le
\E_t\!\left[\sum_{h=1}^{H}
\min\!\left\{1,\frac{P^\pi(\xi_{t,h})}
{P^{\pi_t}(\xi_{t,h})}\right\}
\bigl(e_h(\xi_{t,h};\pi)+e_h(\xi_{t,h};\pi_t)\bigr)\right]    \\ 
&= \E_{\pi_t \sim w_t}\left[
\operatorname{Trun}_{\pi_t}\left[\overline L(\pi)\right]
+\operatorname{Trun}_{\pi}\left[\overline L(\pi_t)\right]\right] 
\leq \overline L(\pi)+ \E_{\pi_t \sim w_t}\left[\overline L(\pi_t)\right]. 
\end{align*}
Weighting  by $w_t(\pi)$ yields $ \E_t\!\left[\sum_{\pi\in\Pi}w_t(\pi)\ell_t(\pi)^2\right]
\le 2\sum_{\pi\in\Pi}w_t(\pi)\overline L(\pi)$. 

Combining everything above with \cref{eq:exp bound} gives
\[
\begin{aligned}
\frac1{64}\E\!\left[\sum_{t=1}^T\sum_{\pi\in\Pi}
w_t(\pi)\overline L(\pi)\right]
\le{}&
2TL^\star+\frac{\log|\Pi|}{\eta} +2\eta\,\E\!\left[\sum_{t=1}^T\sum_{\pi\in\Pi}
w_t(\pi)\overline L(\pi)\right].
\end{aligned}
\]
With $\eta=1/256$ as in \cref{alg:online}, self-bounding yields
\(
\E\big[\frac1T\sum_{t}\sum_{\pi}
w_t(\pi)\overline L(\pi)\big]
\leq O\big( L^\star+\frac{\log|\Pi|}{T}
\big)\), 
which further implies 
\(
\E\big[\frac1T\sum_{t}\sum_{\pi}
w_t(\pi)L(\pi)\big]
\leq O\big(B\big[L^\star+\frac{\log|\Pi|}{T}\big]\big)
\)
by \cref{eq:visible} under \cref{ass:relative-second-moment}.

Since $\pi_t\sim w_t$, this gives an expected-risk bound when the output policy $\widehat{\pi}$ is uniformly sampled from $\{\pi_1, \cdots, \pi_T\}$. Our \cref{alg:online} instead returns $\widehat\pi\in\operatorname*{arg\,max}_{\pi\in\Pi}w_{T+1}(\pi)$ for a high-probability bound. 
Its high-probability analysis uses the same
second-moment bounds that control the quadratic term to bound
accumulated sampling fluctuations. The full high-probability analysis appears in \cref{app:online}.
\end{proof}


\paragraph{Extension to Hellinger distance.} The same algorithmic approach also applies to Hellinger
distance 
\relax{$\Hell^2(P,Q)=\sum_z(\sqrt{P(z)}-\sqrt{Q(z)})^2$.
We define policy $\pi$'s local loss on prefix $\xi_h$ as 
$g_h(\xi_h;\pi)= $} \relax{ $
 \sqrt{\frac{P^{\pi^\star}(\xi_h)}{P^\pi(\xi_h)}}
 \Hell^2\!\left(
 \pi_h(\cdot | s_h),\pi_h^\star(\cdot| s_h)\right)
$ with
$\E_{P^\pi}\big[\sum_{h=1}^H g_h(\xi_h;\pi)\big]
=\Hell^2(P^\pi,P^{\pi^\star})$.
We also clip} the cumulative local loss by letting
$e_h^{\mathrm H}(\xi_h;\pi) = \min\{1,\sum_{u=1}^h g_u(\xi_u;\pi)\} - \min\{1,\sum_{u=1}^{h-1} g_u(\xi_u;\pi)\}$. 
The algorithm then uses losses
\(
 \ell_t(\pi)=
 \sum_{h=1}^H
 \min\big\{1,
 \frac{P^\pi(\xi_{t,h})}{P^{\pi_t}(\xi_{t,h})}\big\}
 \left(
 e_h^{\mathrm H}(\xi_{t,h};\pi)
 -
 e_h^{\mathrm H}(\xi_{t,h};\pi_t)
 \right)
\)
in the same exponential-weights update as \cref{alg:online}. The full algorithm for online Hellinger distance learning is provided in \cref{alg:online-hellinger}. 
\cref{thm:hellinger-bound} establishes constant approximation
and estimation coefficients for online learning with full feedback,
matching the guarantee of \citet{rohatgi2025} for offline learning
from teacher trajectories. 

\begin{theorem}
Let $\widehat{\pi} \in \Pi$ be the output policy of \cref{alg:online-hellinger}. With probability at least $1-\delta$,
\(
 \Hell^2(P^{\widehat\pi},P^{\pi^\star})
 \lesssim\min_{\pi\in\Pi}\Hell^2(P^\pi,P^{\pi^\star})
 + \log(|\Pi|/\delta)/T.
\) 
\label{thm:hellinger-bound}
\end{theorem}

Under \cref{ass:ratio-cap}, every policy $\pi \in \Pi$ satisfies $\Hell^2(P^\pi,P^{\pi^\star})
 \le \KL\left( P^\pi \| P^{\pi^\star} \right)
 \lesssim B_\infty\Hell^2(P^\pi,P^{\pi^\star})$ because $(\sqrt{x}-1)^2
 \le x\log x-x+1
 \lesssim B_\infty(\sqrt{x}-1)^2$ for $0\le x\le e^{B_\infty}$. Thus, the guarantee of Hellinger distance can be transferred to reverse KL divergence at the cost of $B_{\infty}$. Thus, \cref{thm:hellinger-bound} directly implies $L(\widehat\pi)\le O\!\left(B_\infty\left[L^\star + \log(|\Pi|/\delta)/T \right]\right)$. However, a similar reduction does not hold under \cref{ass:relative-second-moment} (see the construction in \cref{prop:offline-action-B-lower}), so \cref{alg:online} provides more general guarantees.

\subsection{The Limitation of Action Feedback in the Online Setting}\label{sec:online-action}

\cref{alg:online} uses the teacher's full action distribution
to evaluate the local KL losses of every $\pi \in \Pi$.
With action feedback, the learner only observes the teacher's probability on its chosen action, while the probabilities of other actions remain unobserved. The following lower bound shows that this
missing information can significantly increase the sample requirement of any learner, even in the realizable setting.

\begin{theorem}[Online action-feedback lower bound]\label{thm:online-token-lower} For a sufficiently large $T$, every online learner returning a policy $\widehat{\pi} \in \Pi$ with action feedback from the teacher  has a realizable instance
  satisfying \cref{ass:ratio-cap}, on which $L(\widehat\pi)=\widetilde\Omega\!\left(
    \min\{|\mathcal A|,|\Pi|,e^{B_\infty}\} \frac{B_\infty \log|\Pi|}{T}\right)$ with probability at least $1/4$.
\end{theorem}

Compared with the full-feedback guarantee in
\cref{thm:online}, this lower bound requires an additional
factor of $\min\{|\mathcal A|,|\Pi|,e^{B_\infty}\}$ in the estimation error even in the realizable setting.
Full feedback therefore provides a statistical advantage over action feedback in the online learning setting.



%% file: sections/offline_learning.tex
\section{Offline learning from teacher trajectories}\label{sec:offline}

In \cref{sec:online}, the learner chooses a rollout policy
in each round and uses the observed losses to guide future rollouts. In the offline setting, the learner instead receives $T$ independent teacher
trajectories $\tau_t^\star\sim P^{\pi^\star}$ with $\tau_t^\star = (s_{t,1}^\star, a_{t,1}^\star, \cdots, s_{t,H}^\star, a_{t,H}^\star)$. We will discuss algorithms to deal with both full feedback and action feedback for the offline setting in this section.
\subsection{Offline learning with full feedback}

Full feedback reveals $k_h(s_{t,h}^\star;\pi)$ for every candidate
$\pi$ at each teacher-visited state. A natural estimator minimizes
the average accumulated local KL over the teacher trajectories,
\begin{equation}\label{eq:offline-unclipped-estimator}
 \widehat\pi_{\rm raw}\in\operatorname*{arg\,min}_{\pi\in\Pi}
 \frac1T\sum_{t=1}^T\sum_{h=1}^H k_h(s_{t,h}^\star;\pi).
\end{equation}
However, \cref{prop:offline-unclipped} shows that although this algorithm can obtain the optimal estimation rate in the realizable setting, it suffers $\Capx=\widetilde{\Omega}(B^2)$ approximation error rate under misspecification even under the stronger pathwise bound $\sum_{h=1}^H k_h(s_h;\pi)\le B$, which is suboptimal compared with the lower bound in \cref{thm:online-optimality}.

\begin{proposition}[Unclipped offline ERM]\label{prop:offline-unclipped}
Under \cref{ass:relative-second-moment}, if $\pi^\star\in\Pi$, then for every
$\delta\in(0,1)$, with probability at least $1-\delta$,
\(
 L(\widehat\pi_{\rm raw})
 \le O\!\left(B\log(|\Pi|/\delta)/T\right)
\).
Under misspecification, even under the stronger pathwise bound $\sum_{h=1}^H k_h(s_h;\pi)\le B$ as in \cref{lem:second-moment-sufficient}, a uniform guarantee for
\cref{eq:offline-unclipped-estimator} with an estimation term
vanishing as $T\to\infty$ requires
$\Capx=\widetilde{\Omega}(B^2)$.
\end{proposition}
The failure under misspecification arises because teacher trajectories may visit states that
a candidate $\pi$ never reaches. Large local KL losses at these
states do not contribute to $L(\pi)$, but can dominate the objective in \cref{eq:offline-unclipped-estimator} and cause it to reject $\pi$ even when
$L(\pi)$ is small. Similar to the online setting, our estimator limits each trajectory's contribution by clipping the
accumulated loss, yielding
\begin{equation}\label{eq:offline-full-estimator}
 \widehat\pi\in\operatorname*{arg\,min}_{\pi\in\Pi}
 \frac1T\sum_{t=1}^T
 \min\!\left\{1,\sum_{h=1}^H k_h(s_{t,h}^\star;\pi)\right\}.
\end{equation}
\begin{theorem}[Offline full-feedback guarantee]\label{thm:offline-full}
The estimator in \cref{eq:offline-full-estimator} satisfies
the high-probability guarantee of \cref{thm:online} under
\cref{ass:relative-second-moment}.
\end{theorem}

Under \cref{ass:relative-second-moment}, clipping preserves the optimal estimation
rate in the realizable setting while improving the approximation
coefficient from the $\widetilde{\Omega}(B^2)$ required by the
unclipped estimator to $O(B)$ in the pathwise special case, matching the lower bound in \cref{thm:online-optimality}.




We now provide a proof sketch for \cref{thm:offline-full}. Let $\widehat\mu(\pi) = \frac1T\sum_{t=1}^T
 \min\big\{1,\sum_{h=1}^H k_h(s_{t,h}^\star;\pi)\big\}$ and $\mu(\pi) = 
\E_{P^{\pi^\star}}\big[
 \min\big\{1,\sum_{h=1}^H k_h(s_{t,h}^\star;\pi)\big\} \big]$. With probability at least $1-\delta$, standard Bernstein concentration for bounded nonnegative losses gives for all
$\pi\in\Pi$,
\[
 \widehat\mu(\pi)\le3\mu(\pi)
 +O\!\left(\frac{\log(|\Pi|/\delta)}T\right),
 \qquad
 \mu(\pi)\le3\widehat\mu(\pi)
 +O\!\left(\frac{\log(|\Pi|/\delta)}T\right).
\]
Since $\widehat\pi$ minimizes $\widehat\mu$ in \cref{eq:offline-full-estimator}, these inequalities give
\[
 \mu(\widehat\pi)
 \le3\widehat\mu(\widehat\pi)
 +O\!\left(\frac{\log(|\Pi|/\delta)}T\right)
 =3\min_{\pi\in\Pi}\widehat\mu(\pi)
 +O\!\left(\frac{\log(|\Pi|/\delta)}T\right)
 \le9\min_{\pi\in\Pi}\mu(\pi)
 +O\!\left(\frac{\log(|\Pi|/\delta)}T\right).
\]
To get the target risk bound on $L(\pi)$, we need to connect $\mu(\pi)$ with $\overline L(\pi)$, and bound $L(\pi)$ through \cref{eq:visible}. The challenge is that $\mu(\pi)$ averages losses
under $P^{\pi^\star}$, whereas $\overline L(\pi)$ averages under
$P^\pi$. After clipping, this change of measure costs only a constant factor. \cref{lem:visibility-main} establishes
\(\mu(\pi)\asymp\overline L(\pi).
\) Thus, with probability
at least $1-\delta$,
\[
\begin{aligned}
 \overline L(\widehat\pi) \lesssim \mu(\widehat\pi)\lesssim \min_{\pi\in\Pi}\mu(\pi)
          +\frac{\log(|\Pi|/\delta)}{T} \lesssim \min_{\pi\in\Pi}\overline L(\pi)
          +\frac{\log(|\Pi|/\delta)}{T} \le L^\star+\frac{\log(|\Pi|/\delta)}{T}.
\end{aligned}
\]
\cref{eq:visible} converts this bound to $L(\widehat\pi)$,
introducing a factor $B$ under \cref{ass:relative-second-moment}.
\cref{app:common,app:offline} give the full
change-of-measure and concentration arguments.

\subsection{Offline learning with action feedback}

Action feedback reveals the teacher's probability only for the
observed action, so the local KL losses in
\cref{eq:offline-full-estimator} are generally unavailable.
Even under realizability, \cref{ass:relative-second-moment} alone does not permit
a uniformly vanishing risk bound for offline learners with action feedback 
(\cref{prop:offline-action-B-lower}).
We instead only consider \cref{ass:ratio-cap} and use the offline Hellinger estimator of
\citet{rohatgi2025}, which returns $\widehat\pi\in\Pi$ satisfying,
with probability at least $1-\delta$,
$\Hell^2(P^{\widehat\pi},P^{\pi^\star})
\lesssim\min_{\pi\in\Pi}\Hell^2(P^\pi,P^{\pi^\star})
+\log(|\Pi|/\delta)/T$.
Under \cref{ass:ratio-cap}, the Hellinger-to-KL conversion in
\cref{sec:onlin alg} gives
$L(\widehat\pi)\lesssim
B_\infty\Hell^2(P^{\widehat\pi},P^{\pi^\star})$ and
$\min_{\pi\in\Pi}\Hell^2(P^\pi,P^{\pi^\star})\le L^\star$.
Substituting these inequalities gives the following guarantee.

\begin{proposition}[Offline action-feedback guarantee]\label{thm:offline-token}
Under \cref{ass:ratio-cap}, let $\widehat\pi\in\Pi$ be the output
of the estimator in Equation~(11) of \citet{rohatgi2025}.
For every $\delta\in(0,1)$, with probability at least $1-\delta$,
\begin{equation}\label{eq:offline-token}
 L(\widehat\pi)\le O\!\left(B_\infty
 \left[L^\star+\frac{\log(|\Pi|/\delta)}{T}\right]\right).
\end{equation}
\end{proposition}
The lower bounds in \cref{thm:online-optimality}
also hold for offline learning with action feedback under \cref{ass:ratio-cap}. Thus, the reduction from Hellinger to reverse KL learning is minimax optimal in this setting. \cref{thm:offline-token} and
\cref{thm:online-token-lower} show that, under
\cref{ass:ratio-cap}, offline action-feedback learning can require
exponentially fewer episodes than online action-feedback learning
in the worst case.

%% file: sections/conclusion.tex
\section{Conclusion}\label{sec:conclusion}
We characterize how data source and teacher feedback shape reverse-KL
learning under misspecification. For the online learning setting, we show the failure of common empirical algorithms and provide the first algorithm achieving the optimal guarantee under full feedback and a relative second-moment condition. We further show that full feedback can significantly improve online sample
efficiency over action feedback. For the offline setting, we also establish algorithms with tight bounds, demonstrating that
under a trajectory likelihood-ratio bound, teacher trajectories can avoid the additional sample cost of
online learning with action feedback.
Future work includes developing and empirically evaluating practical
distillation algorithms based on our prefix reweighting and trajectory
clipping mechanism, and determining the optimal statistical guarantees
achievable by computationally efficient learners.

\clearpage

%% file: appendices/related_work.tex
\clearpage
\section{Related work}\label[appendix]{app:related-work}

\paragraph{Offline imitation learning theory}
\label{app:related-oracle}

Offline imitation learning asks what can be learned from a fixed set
of expert demonstrations without further interaction.
In the realizable setting, classical analyses
of zero-one-loss behavior cloning relate action-prediction errors to
compounding errors in task performance \citep{ross2010}.
\citet{rajaraman2020} characterize minimax task-performance guarantees
in finite-state MDPs, showing that error compounding can be unavoidable
when dynamics are unknown.
With known dynamics and a deterministic expert, improved dependence on
the horizon is possible \citep{rajaraman2021}.
For general policy classes under realizability, \citet{foster2024}
show that log-loss behavior cloning admits horizon-free guarantees
when the cumulative reward range and policy-class complexity are controlled.
Additional offline data provide another source of information:
MILO learns a dynamics model from auxiliary transition data and
mitigates covariate shift when those data sufficiently cover the
expert's behavior \citep{chang2021milo}.
Our offline learners instead use teacher trajectories and the
specified probability feedback, without auxiliary transition data.

When the student policy class is not expressive enough to cover the teacher, the question becomes
how closely a learner can compete with the best available policy.
A common target is an oracle bound of the form
$D(\widehat\pi;\pi^\star)\le
\Capx\min_{\pi\in\Pi}D(\pi;\pi^\star)
+C_{\rm est}\log(|\Pi|/\delta)/T$ where $D(\cdot;\cdot)$ can be any distance measure.
For trajectory-level Hellinger distance, \citet{rohatgi2025} give an
offline estimator with constant approximation and estimation
coefficients, while establishing an $\Omega(H)$ approximation barrier
for a broad class of next-token methods.
\citet{xu2026} study trajectory-level forward KL divergence, obtaining constant-factor
approximation error with $\Theta(HG)$  estimation error where \(
\sup_{\pi\in\Pi}\sup_{h\in[H]}\sup_{s_h}
\log\frac{\pi_h^\star(x_h\mid s_h)}
{\pi_h(x_h\mid s_h)}
\le G.
\) Both \citet{rohatgi2025} and \citet{xu2026} study offline learning with action feedback. In the same setting, we study reverse KL learning and show that
$\Capx,C_{\rm est}=\Theta(B_\infty)$ under \cref{ass:ratio-cap}. Due to the different directions in forward and reverse KL, we assume the teacher can cover any student, whereas \citet{xu2026} assumes any student can cover the teacher. Teacher coverage is more natural for distillation because a limited-capacity student may concentrate on teacher-supported behaviors without
representing the teacher's entire range of behaviors.

\paragraph{Interactive imitation learning}
\label{app:related-limits}

A student may visit states rarely represented in teacher demonstrations,
especially after making a mistake \citep{ross2010}.
DAgger obtains expert supervision on these states and reduces imitation
to no-regret online learning \citep{ross2011}.
\citet{li2024agnostic} give an oracle-efficient agnostic algorithm
with access to exploratory state samples.
Its comparator measures local imitation error on states visited during
training; ours evaluates each candidate under its own trajectory law.
Thus a small online regret alone does not imply our reverse-KL oracle
bound. 

Several works identify conditions under which interaction
improves imitation-learning performance. \citet{espinosadice2025} avoid quadratic error compounding
under a completeness condition on the policy class.
\citet{zhang2026} identify an offline information bottleneck under
misspecification even at horizon one.
\citet{sriraman2026} establish an exponential benefit from interaction
with a noisy expert when performance is measured against a clean expert.
\citet{viano2026} obtain efficient on-policy imitation when the expert's
Q-function, rather than its policy, is realizable.
These guarantees concern different targets and
assumptions from our comparison of online and offline reverse-KL learning rates.

\paragraph{Language-model distillation}
\label{app:related-distillation}\label{app:related-feedback}
\label{app:related-clipping}

Building on knowledge and policy distillation \citep{hinton2015,rusu2016}, language-model distillation can use teacher-generated responses
\citep{kim2016} or student-generated responses
\citep{lin2020,agarwal2024,li2026}, with forward KL and reverse KL as common losses. Beyond the standard losses, \citet{wu2025} propose an adaptive combination of these two losses.
\citet{ko2024} propose skew KL losses that compare a distribution with
a mixture of teacher and student distributions, while \citet{ko2025}
use a contrastive objective to increase the likelihood of teacher responses
and decrease that of student responses.
\citet{gu2024} develop policy-gradient updates for sequence-level
reverse KL, using returns based on accumulated future
teacher-to-student log likelihood ratios. These returns allow later predictions to influence updates
to earlier actions.

The feedback available at each training recipe is a separate choice.
\citet{lu2025} use teacher probabilities for sampled tokens,
\citet{xu2026deepseek} use full next-token distributions, and
\citet{fu2026,yu2026mismatch} incorporate information about
teacher-preferred alternatives.
Our full-versus-action-feedback comparison isolates the statistical
value of the available information by comparing the best attainable
rates under each feedback model.

%% file: appendix_online.tex
\theoremstyle{plain}
\newtheorem*{refittingstatement}{\cref{prop:training-failure}
  (detailed statement)}
\newtheorem*{optimalitystatement}{Lower bounds for
  \cref{thm:online-optimality} and \cref{sec:offline}}
\newtheorem*{actionlowerstatement}{\cref{thm:online-token-lower}
  (detailed statement)}
\theoremstyle{remark}

\section{Online learning}\label{app:online-proofs}\label{app:upper-bounds}
\subsection{Failure of exact empirical fitting}\label{app:training-barriers}
\label[appendix]{app:training-failure}
At round $t$, the current policy $\pi_t$ generates a nonempty batch of
$H$-step trajectories with full teacher feedback. Let $\mathcal D_t$ contain all trajectories collected up to and
including round $t$.
The learner assigns weights $\omega_{t,\tau}\ge0$ with
$\sum_{\tau\in\mathcal D_t}\omega_{t,\tau}>0$ and calculates
\begin{equation}\label{eq:training-objectives}
 \begin{aligned}
 \pi_{t+1}&\in\operatorname*{arg\,min}_{\pi\in\Pi}
 Q_t(\pi;\mathcal D_t),\\
 Q_t(\pi;\mathcal D_t)
 &=\sum_{\tau\in\mathcal D_t}\omega_{t,\tau}\phi_t\!\left(
 \sum_{h=1}^H d_t\!\left(\pi_h(\cdot\mid s_h),
 \pi_h^\star(\cdot\mid s_h)\right)\right).
 \end{aligned}
\end{equation}
The weights may depend on the observed history but are the same for
all candidate policies. Assigning weight zero discards a trajectory, so this
rule includes fitting any nonempty subset of the collected trajectories,
using only the latest batch, or reusing all past data.
If the current policy is already an empirical minimizer, it is retained:
\(
 \pi_t\in\operatorname*{arg\,min}_{\pi\in\Pi}Q_t(\pi;\mathcal D_t) \Longrightarrow \pi_{t+1}=\pi_t.
\)
Thus a change of policy requires a strict decrease in the empirical
objective. The output $\widehat\pi$ is the policy returned by the
final step.

For student and teacher distributions $p,q$, the local loss
$d_t(p,q)$ is finite, nonnegative, zero exactly when $p=q$, and convex
in $q$. The function $\phi_t$ is nondecreasing, with $\phi_t(0)=0$ and
$\phi_t(u)>0$ for $u>0$. With $\phi_t(u)=u$, forward KL uses $d_t(p,q)=\KL(q\|p)$ and has the
same minimizers as the teacher-weighted cross-entropy
$-\sum_a q(a)\log p(a)$ used by \citet{lin2020}.
Reverse KL uses $d_t(p,q)=\KL(p\|q)$, as in the local distillation
objective of \citet{agarwal2024}.
Skew reverse KL uses $d_t(p,q)=\KL(p\|\alpha_t p+(1-\alpha_t)q)$
for $\alpha_t\in[0,1)$ \citep{ko2024}.
These losses are convex in $q$ by joint convexity of KL.
The class also includes mixtures of the two KL directions and
clipping with $\phi_t(u)=\min\{c_t,u\}$ for $c_t>0$.

\begin{refittingstatement}
For every sufficiently large $\lambda$, there is an $H=2$, $|\Pi|=2$
instance satisfying
\cref{ass:relative-second-moment,ass:ratio-cap} with $B=4\lambda$,
$B_\infty=8\lambda$, and $L^\star\lesssim\lambda e^{-\lambda}$.
Every update rule defined in \cref{eq:training-objectives},
initialized uniformly over $\Pi$ and using
$T=\lceil\lambda^3\rceil$ episodes, returns $\widehat\pi$ satisfying
\[
 \Prob\!\left\{L(\widehat\pi)\ge\lambda/4\right\}\ge\frac14.
\]
\end{refittingstatement}

\begin{proof}[Proof of \cref{prop:training-failure}]
\emph{MDP and policy class.}
The two-step MDP has
\[
 \mathcal A=\{a_1,a_2\},\qquad
 \mathcal S_1=\{s_{1,1},s_{1,2}\},\qquad
 \mathcal S_2=\{s_{2,1},s_{2,2}\},
 \qquad \mu(s_{1,1})=\mu(s_{1,2})=\tfrac12.
\]
Its transitions are
\[
 \begin{aligned}
 \mathsf P_1(s_{2,1}\mid s_{1,1},a_1)&=1,\\
 \mathsf P_1(s_{2,2}\mid s,a)&=1
 \quad\bigl((s,a)\in(\mathcal S_1\times\mathcal A)
                   \setminus\{(s_{1,1},a_1)\}\bigr).
 \end{aligned}
\]
The episode ends after the second action.
Set $\epsilon=e^{-\lambda}$, $\bar p=1-\epsilon^2$, and
$v=e^{-4\lambda}$, and let $\Pi=\{\pi_{\rm good},\pi_{\rm bad}\}$.
The table gives the probability of $a_1$ at each state; the
probability of $a_2$ is its complement.
\begin{center}
\begin{tabular}{@{}lcccc@{}}
\toprule
Policy & $s_{1,1}$ & $s_{1,2}$ & $s_{2,1}$ & $s_{2,2}$\\
\midrule
$\pi^\star$ & $\epsilon$ & $\epsilon$ & $v$ & $1/2$\\
$\pi_{\rm good}$ & $\epsilon$ & $\epsilon^2$ & $\bar p$ & $1/2$\\
$\pi_{\rm bad}$ & $\bar p$ & $\epsilon$ & $v$ & $1/2$\\
\bottomrule
\end{tabular}
\end{center}
Write $\Ber(p)=(p,1-p)$ on $(a_1,a_2)$.
All action probabilities are positive.

\emph{Boundedness and trajectory risks.}
The only candidate--state pairs at which the policies differ are
listed below. For $\lambda\ge1$,
\[
 \begin{array}{c|c|c|c}
 \pi&s_h&\displaystyle\max_a\frac{\pi_h(a\mid s_h)}
                              {\pi_h^\star(a\mid s_h)}
       &k_h(s_h;\pi)\\ \hline
 \pi_{\rm bad}&s_{1,1}&\bar p/\epsilon\le e^\lambda&\le\lambda\\
 \pi_{\rm good}&s_{1,2}&(1-\epsilon^2)/(1-\epsilon)=1+\epsilon
       &\le\log(1+\epsilon)\le\epsilon\\
 \pi_{\rm good}&s_{2,1}&\bar p/v\le e^{4\lambda}&\le4\lambda
 \end{array}
\]
A feasible trajectory cannot contain both $s_{1,2}$ and $s_{2,1}$.
For each $\pi\in\Pi$ and $\tau\in\mathcal T_{\rm env}$,
\begin{align*}
 \sum_{h=1}^2 k_h(s_h;\pi)&\le4\lambda \\
 \log\frac{P^\pi(\tau)}{P^{\pi^\star}(\tau)}
 &=\sum_{h=1}^2\log\frac{\pi_h(a_h\mid s_h)}
                               {\pi_h^\star(a_h\mid s_h)}
 \le4\lambda\le8\lambda.
\end{align*}
Thus \cref{ass:relative-second-moment,ass:ratio-cap} hold with the stated budgets, using \cref{lem:second-moment-sufficient} for the first assumption.
The trajectory risks satisfy
\begin{equation}\label{eq:training-risk}
 \begin{aligned}
 L(\pi_{\rm good})
 &=\tfrac12\left[k_1(s_{1,2};\pi_{\rm good})
          +\epsilon k_2(s_{2,1};\pi_{\rm good})\right]
 \le\tfrac12(1+4\lambda)\epsilon,\\
 L(\pi_{\rm bad})
 &=\tfrac12\KL(\Ber(\bar p)\|\Ber(\epsilon))
 \ge\tfrac12(\bar p\lambda-\log2)\ge\lambda/4.
 \end{aligned}
\end{equation}
The first equality in each line is the KL chain rule; the second
line uses binary entropy at most $\log2$ and sufficiently large
$\lambda$. Hence $L^\star=L(\pi_{\rm good})\lesssim\lambda e^{-\lambda}$.

\emph{Empirical loss on each observed trajectory.}
For
$\theta=(\bar p-\epsilon)/(\bar p-v)\in(0,1)$,
$\Ber(\epsilon)=\theta\Ber(v)+(1-\theta)\Ber(\bar p)$.

On a trajectory with $(s_1,s_2)=(s_{1,1},s_{2,1})$,
\begin{align}
 &\sum_{h=1}^2 d_t\!\left(
 \pi_{{\rm bad},h}(\cdot\mid s_h),
 \pi_h^\star(\cdot\mid s_h)\right)
 \notag\\
 &\quad=d_t(\Ber(\bar p),\Ber(\epsilon))
 \tag{$d_t(\Ber(v),\Ber(v))=0$}\\
 &\quad\le
 \theta d_t(\Ber(\bar p),\Ber(v))
 +(1-\theta)d_t(\Ber(\bar p),\Ber(\bar p))
 \tag{Convexity in the second parameter}\\
 &\quad=\theta d_t(\Ber(\bar p),\Ber(v))
 <d_t(\Ber(\bar p),\Ber(v))
 \tag{$d_t(p,p)=0$, $0<\theta<1$, $v\ne\bar p$}\\
 &\quad=\sum_{h=1}^2 d_t\!\left(
 \pi_{{\rm good},h}(\cdot\mid s_h),
 \pi_h^\star(\cdot\mid s_h)\right).
 &&(\text{$d_t(\Ber(\epsilon),\Ber(\epsilon))=0$})
 \label{eq:refitting-first-path}
\end{align}
On a trajectory with $(s_1,s_2)=(s_{1,2},s_{2,2})$,
\begin{align}
 &\sum_{h=1}^2 d_t\!\left(
 \pi_{{\rm bad},h}(\cdot\mid s_h),
 \pi_h^\star(\cdot\mid s_h)\right)=0
 \tag{$\pi_{\rm bad}=\pi^\star$ at both states}\\
 &\quad<d_t(\Ber(\epsilon^2),\Ber(\epsilon))
 \tag{$d_t(p,q)>0$ for $p\ne q$, $\epsilon^2\ne\epsilon$}\\
 &\quad=\sum_{h=1}^2 d_t\!\left(
 \pi_{{\rm good},h}(\cdot\mid s_h),
 \pi_h^\star(\cdot\mid s_h)\right).
 &&(\text{$d_t(\Ber(1/2),\Ber(1/2))=0$})
 \label{eq:refitting-second-path}
\end{align}

Whenever every trajectory in $\mathcal D_t$ follows one of these
two state sequences,
\begin{align}
 Q_t(\pi_{\rm bad};\mathcal D_t)
 &=\sum_{\tau\in\mathcal D_t}\omega_{t,\tau}\phi_t\!\left(
   \sum_{h=1}^2d_t\!\left(\pi_{{\rm bad},h}(\cdot\mid s_h),
                         \pi_h^\star(\cdot\mid s_h)\right)\right)
 \tag{\cref{eq:training-objectives}}\\
 &\le\sum_{\tau\in\mathcal D_t}\omega_{t,\tau}\phi_t\!\left(
   \sum_{h=1}^2d_t\!\left(\pi_{{\rm good},h}(\cdot\mid s_h),
                         \pi_h^\star(\cdot\mid s_h)\right)\right)
 \tag{\cref{eq:refitting-first-path,eq:refitting-second-path},
       $\omega_{t,\tau}\ge0$, $\phi_t$ nondecreasing}\\
 &=Q_t(\pi_{\rm good};\mathcal D_t).
 &&(\text{\cref{eq:training-objectives}})
 \label{eq:refitting-objective-order}
\end{align}
Since the update keeps the current policy whenever it is an
empirical minimizer, \cref{eq:refitting-objective-order} gives
\[
 \pi_t=\pi_{\rm bad}\quad\Longrightarrow\quad\pi_{t+1}=\pi_{\rm bad},
\]
including ties caused by clipping.

\emph{Persistence and failure probability.}
The only remaining state sequence is $s_{1,1}\to s_{2,2}$.
Under $\pi_{\rm bad}$, its probability is
\[
 P^{\pi_{\rm bad}}\{s_1=s_{1,1},s_2=s_{2,2}\}
 =\tfrac12\pi_{{\rm bad},1}(a_2\mid s_{1,1})
 =\frac{\epsilon^2}{2}.
\]
Let $\mathcal E_m$ be the event that $\pi_1=\pi_{\rm bad}$
and none of the first $m$ episodes takes $s_{1,1}\to s_{2,2}$.
On $\mathcal E_{m-1}$, every update keeps $\pi_{\rm bad}$
by \cref{eq:refitting-objective-order}, so episode $m$ also uses
$\pi_{\rm bad}$. Therefore,
\begin{equation}\label{eq:refitting-survival}
 \Prob(\mathcal E_0)=\tfrac12,\qquad
 \Prob(\mathcal E_m\mid\mathcal E_{m-1})=1-\epsilon^2/2
 \quad(m\in[T]),\qquad
 \widehat\pi=\pi_{\rm bad}\ \text{on }\mathcal E_T.
\end{equation}
For arbitrary batch sizes and allowed trajectory weights,
\begin{align*}
 \Prob\!\left\{L(\widehat\pi)\ge\lambda/4\right\}
 &\ge\Prob(\mathcal E_T)
 \tag{\cref{eq:training-risk,eq:refitting-survival}}\\
 &=\frac12\left(1-\frac{\epsilon^2}{2}\right)^T
 \ge\frac12\left(1-\frac{T\epsilon^2}{2}\right)
 \tag{Bernoulli's inequality}\\
 &=\frac12\left(1-\frac{\lceil\lambda^3\rceil e^{-2\lambda}}2\right)
 \ge\frac14.
 \tag{Sufficiently large $\lambda$}
\end{align*}
\end{proof}

\subsection{Online reverse-KL learning}\label[appendix]{app:online}

\subsubsection{Clipping and prefix truncation}\label[appendix]{app:common}

\begin{lemma}[Clipped objectives and truncation]\label{lem:visibility-main}
If $L(\pi)<\infty$ for every candidate, then
\begin{equation}\label{eq:visibility-main}
 \E_{P^{\pi^\star}}\!\left[\sum_{h=1}^H e_h(\xi_h;\pi)\right]
 \asymp\overline L(\pi),
 \qquad
 \Hell^2(P^\pi,P^{\pi^\star})\le6\overline L(\pi).
\end{equation}
For every $\pi,\pi'\in\Pi$,
$0\le\operatorname{Trun}_{\pi'}[\overline L(\pi)]\le\overline L(\pi)$ and
\begin{equation}\label{eq:truncation-hellinger-remainder}
 \operatorname{Trun}_{\pi'}\!\left[\overline L(\pi)\right]
 \ge\frac1{16}\overline L(\pi)
 -\Hell^2(P^{\pi^\star},P^{\pi'}).
\end{equation}
Consequently,
\[
 \overline L(\pi)\le64\left(
 \operatorname{Trun}_{\pi'}\!\left[\overline L(\pi)\right]
 +\overline L(\pi')\right).
\]
\end{lemma}

To prove \cref{lem:visibility-main}, we first introduce several auxiliary lemmas to control the behaviour of the clipped loss.

\begin{lemma}[Change of measure for a bounded loss]
\label{lem:bounded-loss-change-measure}
For probability laws $R,S$ and a measurable $f\in[0,1]$,
$\KL(R\|S)\le\E_R \left[f\right]$ implies $\E_S \left[f\right]\asymp\E_R \left[f\right]$.
\end{lemma}
\begin{proof}
\cref{lem:kl-variational}, applied to $g=\lambda f$, gives
for every $\lambda\in\mathbb R$,
\[
 \KL(R\|S)\ge\lambda\E_R \left[f\right]-\log\E_S \left[e^{\lambda f}\right]
 \ge\lambda\E_R \left[f\right]-(e^\lambda-1)\E_S \left[f\right].
\]
The second inequality uses convexity of the exponential on $[0,1]$.
The choices $\lambda=2$ and $\lambda=-1$, respectively, give
$(e^2-1)\E_S \left[f\right]\ge\E_R \left[f\right]$ and
$(1-e^{-1})\E_S \left[f\right]\le2\E_R \left[f\right]$.
\end{proof}

Each term in $\operatorname{Trun}_{\pi'}[\overline L(\pi)]$ can be
written as
$\min\{e_h(\xi_h;\pi)P^\pi(\xi_h),
e_h(\xi_h;\pi)P^{\pi'}(\xi_h)\}$.
For a fixed candidate $\pi$ and any trajectory distribution $P$,
we therefore define an auxiliary distribution $\Rcal_\pi[P]$ that
assigns probability $e_h(\xi_h;\pi)P(\xi_h)$ to each prefix outcome
$(h,\xi_h)$, $h\in[H]$. The distribution also includes
complete-trajectory outcomes $(\infty,\tau)$ for the remaining probability, and is defined by
\begin{equation}\label{eq:kernel-laws}
 \begin{aligned}
 \Rcal_\pi[P](h,\xi_h)&=e_h(\xi_h;\pi)P(\xi_h),\\
 \Rcal_\pi[P](\infty,\tau)
 &=\left(1-\sum_{h=1}^H e_h(\xi_h;\pi)\right)P(\tau).
 \end{aligned}
\end{equation}
The label $\infty$ distinguishes a complete trajectory from a prefix. The difference between $\tau$ and the last prefix $\xi_H$ is that $\xi_H$ do not have $a_H$ but $\tau$ has. Given $\tau\sim P$, these probabilities correspond to retaining its
prefix $(h,\xi_h)$ with probability $e_h(\xi_h;\pi)$ and retaining
$(\infty,\tau)$ with the remaining probability.
Since $\sum_{h=1}^H e_h(\xi_h;\pi)=\min\{1,\sum_{h=1}^H k_h(s_h;\pi)\}$,
all probabilities are nonnegative and sum to one.

\begin{lemma}[KL between auxiliary distributions]\label{lem:localization}
For $L(\pi)<\infty$,
\begin{equation}\label{eq:localization}
 \KL(\Rcal_\pi[P^\pi]\|\Rcal_\pi[P^{\pi^\star}])
 =\E_{P^\pi}\!\left[\sum_{h=1}^H
 \left(1-\sum_{v=1}^h k_v(s_v;\pi)\right)_+k_h(s_h;\pi)\right]
 \le\overline L(\pi).
\end{equation}
\end{lemma}
\begin{proof}
Fix $\pi\in\Pi$.
For each $\widetilde{\pi}\in\Pi\cup\{\pi^\star\}$, define the joint law
$\Prob_{\widetilde{\pi}}$ by $\tau\sim P^{\widetilde{\pi}}$ and
$\sigma\in[H]\cup\{\infty\}$ with conditional distribution
\[
 \begin{aligned}
 \Prob_{\widetilde{\pi}}(\sigma=h\mid\tau)
 &=e_h(\xi_h;\pi),\qquad h\in[H],\\
 \Prob_{\widetilde{\pi}}(\sigma=\infty\mid\tau)
 &=1-\sum_{h=1}^H e_h(\xi_h;\pi).
 \end{aligned}
\]
By \cref{eq:increments}, $e_h(\xi_h;\pi)\ge0$ and
$\sum_{h=1}^H e_h(\xi_h;\pi)\le1$, so these probabilities sum to one.
For $h\in[H]$,
 \begin{align}
 \Prob_{\widetilde{\pi}}(\sigma\ge h\mid\tau)
 &=1-\sum_{v=1}^{h-1}e_v(\xi_v;\pi),\label{eq: sigma >= h}\\
 \Prob_{\widetilde{\pi}}(\sigma>h\mid\tau)
 &=1-\sum_{v=1}^h e_v(\xi_v;\pi)
 =\left(1-\sum_{v=1}^h k_v(s_v;\pi)\right)_+. \label{eq: sigma > h}
\end{align}
where the second equation uses $1 - \min\{1,x\} = (1-x)_{+}$. Consequently, whenever $\sum_{v=1}^{h-1}e_v(\xi_v;\pi)<1$,
\[
 \Prob_{\widetilde{\pi}}(\sigma=h\mid\sigma\ge h,\tau)
 =\frac{e_h(\xi_h;\pi)}
 {1-\sum_{v=1}^{h-1}e_v(\xi_v;\pi)}
 =\Prob_{\widetilde{\pi}}(\sigma=h\mid\sigma\ge h,\xi_h).
\]
The last equality holds because the ratio depends only on $\xi_h$.

Define the retained record by
\[
 Y=\begin{cases}
 (\sigma,\xi_\sigma),&\sigma\in[H],\\
 (\infty,\tau),&\sigma=\infty.
 \end{cases}
\]
Since $e_h(\xi_h;\pi)$ depends only on $\xi_h$,
\[
 \begin{aligned}
 \Prob_{\widetilde{\pi}}(Y=(h,\xi_h))
 &=P^{\widetilde{\pi}}(\xi_h)\Prob_{\widetilde{\pi}}(\sigma=h\mid\xi_h)\\
 &=P^{\widetilde{\pi}}(\xi_h)e_h(\xi_h;\pi)
 =\Rcal_\pi[P^{\widetilde{\pi}}](h,\xi_h),\\
 \Prob_{\widetilde{\pi}}(Y=(\infty,\tau))
 &=P^{\widetilde{\pi}}(\tau)\left(1-\sum_{h=1}^H e_h(\xi_h;\pi)\right)
 =\Rcal_\pi[P^{\widetilde{\pi}}](\infty,\tau).
 \end{aligned}
\]
Thus $Y\sim\Rcal_\pi[P^{\widetilde{\pi}}]$. From \cref{eq:kernel-laws}, for any outcome with $\Rcal_\pi[P^\pi](Y)>0$, the common
stopping factor cancels from the likelihood ratio. When
$\sigma\in[H]$,
\[
\frac{\Rcal_\pi[P^\pi](Y)}
     {\Rcal_\pi[P^{\pi^\star}](Y)}
=
\frac{e_\sigma(\xi_\sigma;\pi)P^\pi(\xi_\sigma)}
     {e_\sigma(\xi_\sigma;\pi)P^{\pi^\star}(\xi_\sigma)}
=
\frac{P^\pi(\xi_\sigma)}{P^{\pi^\star}(\xi_\sigma)}.
\]
When $\sigma=\infty$, the common factor
$1-\sum_{h=1}^H e_h(\xi_h;\pi)$ cancels in the same way,
leaving $P^\pi(\tau)/P^{\pi^\star}(\tau)$. The two policies share the initial-state distribution and
environment transitions, so
\[
\log\frac{\Rcal_\pi[P^\pi](Y)}
         {\Rcal_\pi[P^{\pi^\star}](Y)}
=
\sum_{h=1}^H\1\{\sigma>h\}
\log\frac{\pi_h(a_h\mid s_h)}
         {\pi_h^\star(a_h\mid s_h)}.
\]
Thus, we have 
\begin{align*}
 &\KL(\Rcal_\pi[P^\pi]\|\Rcal_\pi[P^{\pi^\star}])\\
 &\quad=\E_{\Prob_\pi}\!\left[
 \sum_{h=1}^H\1\{\sigma>h\}
 \log\frac{\pi_h(a_h\mid s_h)}{\pi_h^\star(a_h\mid s_h)}\right]\\
 &\quad=\E_{P^\pi}\!\left[
 \sum_{h=1}^H\Prob_\pi(\sigma>h\mid\tau)
 \log\frac{\pi_h(a_h\mid s_h)}{\pi_h^\star(a_h\mid s_h)}\right]\\
 &\quad=\E_{P^\pi}\!\left[
 \sum_{h=1}^H\left(1-\sum_{v=1}^h k_v(s_v;\pi)\right)_+
 \log\frac{\pi_h(a_h\mid s_h)}{\pi_h^\star(a_h\mid s_h)}\right] \tag{\cref{eq: sigma > h}} \\ 
 &\quad=\E_{P^\pi}\!\left[
 \sum_{h=1}^H\left(1-\sum_{v=1}^h k_v(s_v;\pi)\right)_+
 k_h(s_h;\pi)\right].
\end{align*}
By \cref{eq:increments},
$e_h(\xi_h;\pi)=k_h(s_h;\pi)$ when
$\sum_{v=1}^h k_v(s_v;\pi)<1$, otherwise,
$(1-\sum_{v=1}^h k_v(s_v;\pi))_+=0$. Thus,
\[
 \begin{aligned}
 \left(1-\sum_{v=1}^h k_v(s_v;\pi)\right)_+k_h(s_h;\pi)
 &\le\1\!\left\{\sum_{v=1}^h k_v(s_v;\pi)<1\right\}k_h(s_h;\pi)\\
 &=\1\!\left\{\sum_{v=1}^h k_v(s_v;\pi)<1\right\}e_h(\xi_h;\pi)
 \le e_h(\xi_h;\pi).
 \end{aligned}
\]
Hence
\[
 \KL(\Rcal_\pi[P^\pi]\|\Rcal_\pi[P^{\pi^\star}])
 \le\E_{P^\pi}\!\left[\sum_{h=1}^H e_h(\xi_h;\pi)\right]
 =\overline L(\pi).
\]
The construction also permits infinite local KL under a source $\widetilde{\pi}$,
because
\(
 k_h(s_h;\pi)=\infty, \sum_{v=1}^h e_v(\xi_v;\pi)=1, \Prob_{\widetilde{\pi}}(\sigma>h\mid\tau)=0.
\).
\end{proof}

\begin{lemma}[Shared mass on an event]\label{lem:localized-overlap}
Let $R,S$ be probability distributions on a countable space
$\mathcal X$ with $\KL(R\|S)<\infty$, and let $D\subseteq\mathcal X$.
Write $z=R(D)$ and
$v=\sum_{x\in D}\min\{R(x),S(x)\}$.
For $z>0$,
\begin{equation}\label{eq:localized-overlap}
 \KL(R\|S)\ge z\log(z/v)-z.
\end{equation}
Consequently, $\KL(R\|S)\le z$ implies $v\gtrsim z$.
\end{lemma}
\begin{proof}
For $z=0$, $v=0$. For $z>0$,  $0<v\le z$.
Define
\[
 g(x)=
 \begin{cases}
 \displaystyle\log\frac{R(x)}{\min\{R(x),S(x)\}},
   &x\in D,\ R(x)>0,\\
 0,&\text{otherwise}.
 \end{cases}
\]
Then
\[
 \begin{aligned}
 \E_S\left[e^g\right]
 &=1+\sum_{\substack{x\in D\\R(x)>S(x)}}(R(x)-S(x))\\
 &=1+\sum_{x\in D}\bigl(R(x)-\min\{R(x),S(x)\}\bigr)
 =1+z-v,\\
 \E_R\left[g\right]
 &=-z\sum_{\substack{x\in D\\R(x)>0}}\frac{R(x)}z
 \log\frac{\min\{R(x),S(x)\}}{R(x)}\\
 &\ge
 -z\log\left(\sum_{\substack{x\in D\\R(x)>0}}
 \frac{R(x)}z\frac{\min\{R(x),S(x)\}}{R(x)}\right)
 =z\log(z/v).
 \end{aligned}
\]
The inequality follows from Jensen's inequality.
Since $\E_S\left[e^g\right]=1+z-v\in[1,2]$,
\cref{lem:kl-variational} gives
\[
 \begin{aligned}
 \KL(R\|S)
 &\ge\E_R\left[g\right]-\log\E_S\left[e^g\right]\\
 &\ge z\log(z/v)-\log(1+z-v)\\
 &\ge z\log(z/v)-(z-v)
 \ge z\log(z/v)-z.
 \end{aligned}
\]
Consequently,
\(
 \KL(R\|S)\le z \Longrightarrow 
 z\log(z/v)-z\le z \Longrightarrow 
 v\ge e^{-2}z.
\)
\end{proof}

\begin{proof}[Proof of \cref{lem:visibility-main}]
Fix $\pi\in\Pi$, and let $D=\{(h,\xi_h):h\in[H]\}$ be the set
of prefix outcomes in \cref{eq:kernel-laws}.
For every $\widetilde\pi\in\Pi\cup\{\pi^\star\}$,
\cref{eq:kernel-laws} gives
\[
 \begin{aligned}
 \Rcal_\pi[P^{\widetilde\pi}](D) =\sum_{h=1}^H\sum_{\xi_h}
 e_h(\xi_h;\pi)P^{\widetilde\pi}(\xi_h) =\E_{P^{\widetilde\pi}}\!\left[
 \sum_{h=1}^H e_h(\xi_h;\pi)\right].
 \end{aligned}
\]
\cref{lem:localization} therefore implies
\[
 \KL(\Rcal_\pi[P^\pi]\|\Rcal_\pi[P^{\pi^\star}])
 \le\overline L(\pi)=\Rcal_\pi[P^\pi](D).
\]
Applying \cref{lem:bounded-loss-change-measure} to $\Rcal_\pi[P^\pi]$ and $\Rcal_\pi[P^{\pi^\star}]$ and $f(y) = \1\{y \in D\}$ as the indicator of $D$ yields
 \begin{align}
 \E_{P^{\pi^\star}}\!\left[\sum_{h=1}^H e_h(\xi_h;\pi)\right] =\Rcal_\pi[P^{\pi^\star}](D)\asymp\Rcal_\pi[P^\pi](D)=\overline L(\pi). \label{eq:same order}
 \end{align}

Now we control the trajectory Hellinger distance.  
\begin{align}
 \Hell^2(P^\pi,P^{\pi^\star}) &=  \sum_{\tau}\left(1 - \sum_{h=1}^H e_h(\xi_h;\pi)\right)
 \left(\sqrt{P^\pi(\tau)}-\sqrt{P^{\pi^\star}(\tau)}\right)^2 \nonumber\\&\quad + \sum_{\tau}\left(\sum_{h=1}^H e_h(\xi_h;\pi)\right)
 \left(\sqrt{P^\pi(\tau)}-\sqrt{P^{\pi^\star}(\tau)}\right)^2 \nonumber
 \\&=\sum_{\tau}\left(
 \sqrt{\Rcal_\pi[P^\pi](\infty,\tau)}
 -\sqrt{\Rcal_\pi[P^{\pi^\star}](\infty,\tau)}\right)^2 \nonumber\\&\quad +\sum_{\tau}\left(\sum_{h=1}^H e_h(\xi_h;\pi)\right)
 \left(\sqrt{P^\pi(\tau)}-\sqrt{P^{\pi^\star}(\tau)}\right)^2 \tag{\cref{eq:kernel-laws}} \nonumber
 \\&\le\Hell^2(\Rcal_\pi[P^\pi],\Rcal_\pi[P^{\pi^\star}])
 +\overline L(\pi)
 +\E_{P^{\pi^\star}}\!\left[\sum_{h=1}^H e_h(\xi_h;\pi)\right] \tag{$(\sqrt p-\sqrt q)^2\le p+q$} \nonumber
 \\&\le
 \KL(\Rcal_\pi[P^\pi]\|\Rcal_\pi[P^{\pi^\star}])
 +\overline L(\pi)
 +\E_{P^{\pi^\star}}\!\left[\sum_{h=1}^H e_h(\xi_h;\pi)\right] \nonumber
 \\&\le
 2\overline L(\pi)
 +\E_{P^{\pi^\star}}\!\left[\sum_{h=1}^H e_h(\xi_h;\pi)\right]
 \le6\overline L(\pi). \label{eq:Hellinger-bar-L}
\end{align}
The last line uses \cref{lem:localization} and the explicit bound
$\E_{P^{\pi^\star}}\!\left[\sum_{h=1}^H e_h(\xi_h;\pi)\right]
\le\frac{2}{1-e^{-1}}\overline L(\pi)\le4\overline L(\pi)$
from the proof of \cref{lem:bounded-loss-change-measure}.

For the same event $D=\{(h,\xi_h):h\in[H]\}$, \cref{eq:kernel-laws} gives
\[
 \begin{aligned}
 \sum_{z\in D}\min\{\Rcal_\pi[P^\pi](z),\Rcal_\pi[P^{\pi'}](z)\}
 &=\sum_{h=1}^H\sum_{\xi_h}e_h(\xi_h;\pi)
 \min\{P^\pi(\xi_h),P^{\pi'}(\xi_h)\}\\
 &=\operatorname{Trun}_{\pi'}\left[\overline L(\pi)\right]
 \in[0,\overline L(\pi)].
 \end{aligned}
\]
The interval bound uses
$\min\{P^\pi(\xi_h),P^{\pi'}(\xi_h)\}\le P^\pi(\xi_h)$.
By \cref{eq:localization}, \cref{lem:localized-overlap}
applied to $D$ gives
\[
 \sum_{z\in D}\min\{\Rcal_\pi[P^\pi](z),
 \Rcal_\pi[P^{\pi^\star}](z)\}
 \ge e^{-2}\overline L(\pi).
\]
\cref{eq:auxiliary-hellinger-contraction} and
\cref{eq:overlap-triangle} in \cref{lem:hellinger} give
\begin{align*}
 e^{-1}\sqrt{\overline L(\pi)}
 &\le\sqrt{\sum_{z\in D}\min\{\Rcal_\pi[P^\pi](z),
 \Rcal_\pi[P^{\pi^\star}](z)\}}\\
 &\le\sqrt{\sum_{z\in D}\min\{\Rcal_\pi[P^\pi](z),
 \Rcal_\pi[P^{\pi'}](z)\}}
 +\Hell(\Rcal_\pi[P^{\pi^\star}],\Rcal_\pi[P^{\pi'}])\\
 &\le\sqrt{\operatorname{Trun}_{\pi'}\left[\overline L(\pi)\right]}
 +\Hell(P^{\pi^\star},P^{\pi'}).
\end{align*}
Squaring and using $(x+y)^2\le2x^2+2y^2$ and $e^{-2}>1/8$ gives
\[
 \begin{aligned}
 e^{-2}\overline L(\pi)
 &\le2\operatorname{Trun}_{\pi'}\left[\overline L(\pi)\right]
 +2\Hell^2(P^{\pi^\star},P^{\pi'}),\\
 \operatorname{Trun}_{\pi'}\left[\overline L(\pi)\right]
 &\ge\frac1{16}\overline L(\pi)
 -\Hell^2(P^{\pi^\star},P^{\pi'}),\\
 \overline L(\pi)
 &\le e^2\left(
 \sqrt{\operatorname{Trun}_{\pi'}\left[\overline L(\pi)\right]}
 +\sqrt{6\overline L(\pi')}\right)^2\\
 &\le7e^2\left(
 \operatorname{Trun}_{\pi'}\left[\overline L(\pi)\right]
 +\overline L(\pi')\right)\\
 &\le64\left(
 \operatorname{Trun}_{\pi'}\left[\overline L(\pi)\right]
 +\overline L(\pi')\right).
 \end{aligned}
\]
The last three inequalities use \cref{eq:Hellinger-bar-L},
Cauchy--Schwarz, and $7e^2<64$, respectively.
\end{proof}

\subsubsection{Exponential-weights analysis}\label[appendix]{app:online-ew}
Throughout this subsection, $\E_t$ conditions on the history before
episode $t$ and averages over both $\pi_t\sim w_t$ and its trajectory.
For uniform initialization, $\ell_t(\pi)\in[-1,1]$, and
$0<\eta\le\log 2$, standard exponential-weights analysis
\citep[Eqs.~(27.2)--(27.3)]{lattimore2020bandit} gives
\begin{equation}\label{eq:ew-standard}
 -\sum_{t=1}^T\ell_t(\pi^\dagger)
 \le\frac{\log|\Pi|}{\eta}
 -\sum_{t=1}^T\sum_{\pi\in\Pi}w_t(\pi)\ell_t(\pi)
 +\eta\sum_{t=1}^T\sum_{\pi\in\Pi}w_t(\pi)\ell_t(\pi)^2
\end{equation}
for every fixed $\pi^\dagger\in\Pi$ and every realized loss sequence.
The following lemma relates the loss terms to $\overline L$.

\begin{lemma}[Conditional loss moments]\label{lem:weight-moments}
For \cref{alg:online}, $|\ell_t(\pi)|\le1$ and
\begin{equation}\label{eq:weighted-loss-zero}
 \E_t\left[\sum_{\pi\in\Pi}w_t(\pi)\ell_t(\pi)\right]=0.
\end{equation}
With $a=1/16$ and a numerical $C>0$,
\begin{equation}\label{eq:loss-risk-drift}
 \begin{aligned}
 -\E_t\left[\ell_t(\pi)\right]
 &\ge a\sum_{\pi'\in\Pi}w_t(\pi')\overline L(\pi')-2L(\pi),\\
 \E_t\left[\ell_t(\pi)\right]
 &\ge a\overline L(\pi)
 -C\sum_{\pi'\in\Pi}w_t(\pi')\overline L(\pi').
 \end{aligned}
\end{equation}
The second moments satisfy
\begin{equation}\label{eq:loss-moments}
 \begin{aligned}
 \E_t\left[\ell_t(\pi)^2\right]
 &\le\overline L(\pi)+\sum_{\pi'\in\Pi}w_t(\pi')\overline L(\pi'),\\
 \E_t\left[\sum_{\pi\in\Pi}w_t(\pi)\ell_t(\pi)^2\right]
 &\le2\sum_{\pi'\in\Pi}w_t(\pi')\overline L(\pi').
 \end{aligned}
\end{equation}
\end{lemma}
\begin{proof}
The definition of $\ell_t$ gives the conditional mean and cancellation
\begin{equation}\label{eq:online-loss-mean}
 \begin{aligned}
 \E_t\left[\ell_t(\pi)\mid\pi_t=\pi'\right]
 &=\operatorname{Trun}_{\pi'}\left[\overline L(\pi)\right]
 -\operatorname{Trun}_{\pi}\left[\overline L(\pi')\right],\\
 \E_t\left[\sum_{\pi\in\Pi}w_t(\pi)\ell_t(\pi)\right]
 &=\sum_{\pi,\pi'\in\Pi}w_t(\pi)w_t(\pi')
 \Bigl(\operatorname{Trun}_{\pi'}\left[\overline L(\pi)\right]
 -\operatorname{Trun}_{\pi}\left[\overline L(\pi')\right]\Bigr)
 =0.
 \end{aligned}
\end{equation}
The truncation bound in \cref{lem:visibility-main}, together with
$\overline L(\pi)\le L(\pi)$ and
$\Hell^2(P^\pi,P^{\pi^\star})\le L(\pi)$, yields
\[
 \begin{aligned}
 -\E_t\left[\ell_t(\pi)\mid\pi_t=\pi'\right]
 &\ge a\overline L(\pi')
 -\Hell^2(P^\pi,P^{\pi^\star})-\overline L(\pi)
 \ge a\overline L(\pi')-2L(\pi),\\
 \E_t\left[\ell_t(\pi)\mid\pi_t=\pi'\right]
 &\ge a\overline L(\pi)
 -\Hell^2(P^{\pi'},P^{\pi^\star})-\overline L(\pi')
 \ge a\overline L(\pi)-C\overline L(\pi').
 \end{aligned}
\]
Here the last inequality uses
$\Hell^2(P^{\pi'},P^{\pi^\star})\lesssim\overline L(\pi')$ from the
same lemma. Averaging over $\pi'\sim w_t$ gives the two risk bounds.

Each weighted sum in \cref{eq:online-loss} lies in $[0,1]$, so
$|\ell_t(\pi)|\le1$. The inequality $(x-y)^2\le x+y$ for
$x,y\in[0,1]$ gives the remaining bounds in succession,
\[
 \begin{aligned}
 \E_t\left[\ell_t(\pi)^2\mid\pi_t=\pi'\right]
 &\le\operatorname{Trun}_{\pi'}\left[\overline L(\pi)\right]
 +\operatorname{Trun}_{\pi}\left[\overline L(\pi')\right]
 \le\overline L(\pi)+\overline L(\pi'),\\
 \E_t\left[\ell_t(\pi)^2\right]
 &\le\overline L(\pi)+\sum_{\pi'\in\Pi}w_t(\pi')\overline L(\pi'),\\
 \E_t\left[\sum_{\pi\in\Pi}w_t(\pi)\ell_t(\pi)^2\right]
 &\le\sum_{\pi,\pi'\in\Pi}w_t(\pi)w_t(\pi')
 \bigl(\overline L(\pi)+\overline L(\pi')\bigr)
 =2\sum_{\pi\in\Pi}w_t(\pi)\overline L(\pi).
 \end{aligned}
\]
\end{proof}

\begin{proof}[Proof of \cref{thm:online}]
The case $|\Pi|=1$ is immediate. For $|\Pi|\ge2$, fix
$\pi^\dagger\in\arg\min_{\pi\in\Pi}L(\pi)$ and write
$S_T=\sum_{t=1}^T\sum_{\pi\in\Pi}w_t(\pi)\overline L(\pi)$.
The cumulative weighted clipped risk $S_T$ is random because $w_t$
depends on the training history.

\smallskip\noindent\emph{Cumulative training risk.}
The first inequality in \cref{eq:loss-risk-drift}, summed for
$\pi^\dagger$, bounds $aS_T$ by
$2TL^\star-\sum_{t=1}^T\E_t\left[\ell_t(\pi^\dagger)\right]$.
Separating the observed losses from their conditional expectations
and applying \cref{eq:ew-standard} gives
\begin{equation}\label{eq:comparator-training-risk}
 \begin{aligned}
 aS_T
 &\le2TL^\star-\sum_{t=1}^T\E_t\left[\ell_t(\pi^\dagger)\right]\\
 &=2TL^\star-\sum_{t=1}^T\ell_t(\pi^\dagger)
 +\sum_{t=1}^T\Bigl(\ell_t(\pi^\dagger)
 -\E_t\left[\ell_t(\pi^\dagger)\right]\Bigr)\\
 &\le2TL^\star+\frac{\log|\Pi|}\eta
 +\underbrace{\left(-\sum_{t=1}^T\sum_{\pi\in\Pi}
 w_t(\pi)\ell_t(\pi)\right)}_{\text{weighted loss}}
 +\underbrace{\eta\sum_{t=1}^T\sum_{\pi\in\Pi}
 w_t(\pi)\ell_t(\pi)^2}_{\text{quadratic term}}\\
 &\qquad+\underbrace{\sum_{t=1}^T\Bigl(\ell_t(\pi^\dagger)
 -\E_t\left[\ell_t(\pi^\dagger)\right]\Bigr)}
 _{\text{comparator sampling error}}.
 \end{aligned}
\end{equation}
For any fixed $0<\lambda\le1/2$, the following bounds hold
simultaneously with probability at least $1-\delta/2$, by
\cref{lem:fixed-time} and a union bound.

For the weighted loss,
\begin{align}
 &\left|\sum_{t=1}^T\sum_{\pi\in\Pi}w_t(\pi)\ell_t(\pi)\right|
 \notag\\
 &\quad\le\lambda\sum_{t=1}^T\E_t\left[
 \left(\sum_{\pi\in\Pi}w_t(\pi)\ell_t(\pi)\right)^2\right]
 +\frac{\log(8/\delta)}\lambda
 \tag{\cref{eq:exp-upper,eq:weighted-loss-zero}}\\
 &\quad\le\lambda\sum_{t=1}^T\E_t\left[
 \sum_{\pi\in\Pi}w_t(\pi)\ell_t(\pi)^2\right]
 +\frac{\log(8/\delta)}\lambda
 \tag{Jensen's inequality}\\
 &\quad\le2\lambda S_T+\frac{\log(8/\delta)}\lambda.
 &&(\text{\cref{eq:loss-moments}})
 \label{eq:weighted-loss-concentration}
\end{align}
For the quadratic term, \cref{lem:weight-moments} gives
$0\le\sum_{\pi\in\Pi}w_t(\pi)\ell_t(\pi)^2\le1$, so
\begin{align}
 &\eta\sum_{t=1}^T\sum_{\pi\in\Pi}w_t(\pi)\ell_t(\pi)^2
 \notag\\
 &\quad\le\begin{aligned}[t]
 &\eta\sum_{t=1}^T\E_t\left[
 \sum_{\pi\in\Pi}w_t(\pi)\ell_t(\pi)^2\right]
 +\eta\frac{\log(8/\delta)}\lambda\\
 &\quad+\eta\lambda\sum_{t=1}^T\E_t\left[
 \left(\sum_{\pi\in\Pi}w_t(\pi)\ell_t(\pi)^2\right)^2\right]
 \end{aligned}
 \tag{\cref{eq:exp-upper}}\\
 &\quad\le\eta(1+\lambda)\sum_{t=1}^T\E_t\left[
 \sum_{\pi\in\Pi}w_t(\pi)\ell_t(\pi)^2\right]
 +\eta\frac{\log(8/\delta)}\lambda
 \tag{$x^2\le x$ for $x\in[0,1]$}\\
 &\quad\le2\eta(1+\lambda)S_T
 +\eta\frac{\log(8/\delta)}\lambda.
 &&(\text{\cref{eq:loss-moments}})
 \label{eq:quadratic-concentration}
\end{align}
For the comparator sampling error,
\begin{align}
 &\sum_{t=1}^T\Bigl(\ell_t(\pi^\dagger)
 -\E_t\left[\ell_t(\pi^\dagger)\right]\Bigr)
 \notag\\
 &\quad\le\lambda\sum_{t=1}^T\E_t\left[
 \ell_t(\pi^\dagger)^2\right]+\frac{\log(8/\delta)}\lambda
 \tag{\cref{eq:exp-upper}}\\
 &\quad\le\lambda\bigl(T\overline L(\pi^\dagger)+S_T\bigr)
 +\frac{\log(8/\delta)}\lambda
 \tag{\cref{eq:loss-moments}}\\
 &\quad\le\lambda(TL^\star+S_T)+\frac{\log(8/\delta)}\lambda.
 &&(\text{$\overline L(\pi^\dagger)\le L^\star$})
 \label{eq:comparator-concentration}
\end{align}
Substituting \cref{eq:weighted-loss-concentration},
\cref{eq:quadratic-concentration}, and
\cref{eq:comparator-concentration} for the weighted loss, quadratic
term, and comparator sampling error in
\cref{eq:comparator-training-risk}, respectively, gives
\begin{equation}\label{eq:accumulated-score}
 \begin{aligned}
 aS_T
 &\le2TL^\star+\frac{\log|\Pi|}\eta
 +2\lambda S_T+\frac{\log(8/\delta)}\lambda\\
 &\qquad+2\eta(1+\lambda)S_T
 +\eta\frac{\log(8/\delta)}\lambda
 +\lambda(TL^\star+S_T)+\frac{\log(8/\delta)}\lambda,\\
 \bigl[a-3\lambda-2\eta(1+\lambda)\bigr]S_T
 &\le(2+\lambda)TL^\star+\frac{\log|\Pi|}\eta
 +(2+\eta)\frac{\log(8/\delta)}\lambda.
 \end{aligned}
\end{equation}
For $\eta=1/256$, $a-2\eta=7/128>0$. A sufficiently small fixed
$\lambda>0$ therefore makes the coefficient on $S_T$ positive,
yielding $S_T\lesssim TL^\star+\log(|\Pi|/\delta)$.

\smallskip\noindent\emph{Selection by final weight.}
The maximum-weight output satisfies
$w_{T+1}(\widehat\pi)\ge1/|\Pi|=w_1(\widehat\pi)$.
For every candidate with $w_{T+1}(\pi)\ge w_1(\pi)$, the weight
update and Jensen's inequality bound its cumulative observed loss
by the weighted loss,
\begin{equation}\label{eq:log-weight-change}
 \begin{aligned}
 \sum_{t=1}^T\ell_t(\pi)
 &=-\frac1\eta\log\frac{w_{T+1}(\pi)}{w_1(\pi)}
 -\frac1\eta\sum_{t=1}^T\log\left(
 \sum_{\pi'\in\Pi}w_t(\pi')e^{-\eta\ell_t(\pi')}\right)\\
 &\le-\frac1\eta\log\frac{w_{T+1}(\pi)}{w_1(\pi)}
 +\sum_{t=1}^T\sum_{\pi'\in\Pi}w_t(\pi')\ell_t(\pi')\\
 &\le\sum_{t=1}^T\sum_{\pi'\in\Pi}w_t(\pi')\ell_t(\pi').
 \end{aligned}
\end{equation}
The last inequality uses $w_{T+1}(\pi)\ge w_1(\pi)$.
To convert this observed-loss bound to a risk bound, we apply
\cref{eq:exp-lower} with failure probability $\delta/(2|\Pi|)$
to each candidate and take a union bound over $\Pi$.
The uniform bound over $\Pi$ and the preceding concentration bounds
hold jointly with probability at least $1-\delta$.
On this event, every candidate with $w_{T+1}(\pi)\ge1/|\Pi|$ satisfies
\begin{align*}
 &aT\overline L(\pi) \le \sum_{t=1}^T\E_t\left[\ell_t(\pi)\right]+CS_T
 \tag{\cref{eq:loss-risk-drift}}\\
 &\quad\le\sum_{t=1}^T\ell_t(\pi)
 +\lambda\sum_{t=1}^T\E_t\left[\ell_t(\pi)^2\right]
 +CS_T+\frac{\log(2|\Pi|/\delta)}\lambda
 \tag{\cref{eq:exp-lower}}\\
 &\quad\le\sum_{t=1}^T\ell_t(\pi)
 +\lambda T\overline L(\pi)+(C+\lambda)S_T
 +\frac{\log(2|\Pi|/\delta)}\lambda
 \tag{\cref{eq:loss-moments}}\\
 &\quad\le\sum_{t=1}^T\sum_{\pi'\in\Pi}w_t(\pi')\ell_t(\pi')
 +\lambda T\overline L(\pi)+(C+\lambda)S_T
 +\frac{\log(2|\Pi|/\delta)}\lambda
 \tag{\cref{eq:log-weight-change}}\\
 &\quad\le\lambda T\overline L(\pi)+(C+3\lambda)S_T
 +\frac{\log(2|\Pi|/\delta)+\log(8/\delta)}\lambda.
 \tag{\cref{eq:weighted-loss-concentration}}
\end{align*}
Rearranging with $\lambda<a$ and using
\cref{eq:accumulated-score} gives, simultaneously for every
$\pi$ with $w_{T+1}(\pi)\ge1/|\Pi|$,
\begin{equation}\label{eq:final-clipped-risk}
 \begin{aligned}
 \overline L(\pi)
 &\le\frac{(C+3\lambda)S_T
 +\lambda^{-1}\bigl(\log(2|\Pi|/\delta)+\log(8/\delta)\bigr)}
 {(a-\lambda)T}\\
 &\lesssim L^\star+\frac{\log(|\Pi|/\delta)}T.
 \end{aligned}
\end{equation}
In particular, $\widehat\pi$ satisfies this bound.
Under \cref{ass:relative-second-moment},
\cref{lem:second-moment-clipping} then yields
\[
 L(\widehat\pi)
 \le(1+B)\overline L(\widehat\pi)
 \lesssim B\left(L^\star+\frac{\log(|\Pi|/\delta)}T\right).
\]
\end{proof}

\subsection{Online Hellinger learning from full feedback}\label{app:online-hellinger}
\cref{alg:online-hellinger} replaces the local-KL score in
\cref{alg:online} by a prefix-weighted local Hellinger score
before clipping.

\begin{algorithm}[H]
  \caption{Online Hellinger learning with full teacher feedback}
  \label{alg:online-hellinger}
  \small
  \begin{algorithmic}[1]
    \Require Finite candidate class $\Pi$, episode budget $T$.
    \State Set $w_1(\pi)=1/|\Pi|$ for every $\pi\in\Pi$ and $\eta=1/128$.
    \For{$t=1,\ldots,T$}
      \State Sample a policy $\pi_t\sim w_t$ and generate a complete
        trajectory $\tau_t=(s_{t,1},a_{t,1},\ldots,s_{t,H},a_{t,H})
        \sim P^{\pi_t}$ by executing $\pi_t$ for the entire episode.
      \State Observe $\pi_h^\star(\cdot\mid s_{t,h})$ for every
        $h=1,\ldots,H$. Write
        $\xi_{t,h}=(s_{t,1},a_{t,1},\ldots,s_{t,h})$.
      \For{each $\pi\in\Pi$}
        \State For every $h=1,\ldots,H$ with
          $P^\pi(\xi_{t,h})>0$, compute
          \[
            g_h(\xi_{t,h};\pi)=
            \sqrt{\prod_{v=1}^{h-1}
            \frac{\pi_v^\star(a_{t,v}\mid s_{t,v})}
            {\pi_v(a_{t,v}\mid s_{t,v})}}
            \Hell^2\!\left(\pi_h(\cdot\mid s_{t,h}),
            \pi_h^\star(\cdot\mid s_{t,h})\right).
          \]
          Set $g_h(\xi_{t,h};\pi)=0$ when
          $P^\pi(\xi_{t,h})=0$. The empty product equals one.
        \State For every $h=1,\ldots,H$, compute the clipped increment
          \[
            e_h^{\mathrm H}(\xi_{t,h};\pi)=
            \min\!\left\{1,\sum_{v=1}^{h}g_v(\xi_{t,v};\pi)\right\}
            -\min\!\left\{1,\sum_{v=1}^{h-1}g_v(\xi_{t,v};\pi)\right\}.
          \]
      \EndFor
      \State For every $\pi\in\Pi$, compute
        \[
          \ell_t(\pi)=\sum_{h=1}^H
          \min\!\left\{1,\prod_{v=1}^{h-1}
          \frac{\pi_v(a_{t,v}\mid s_{t,v})}
          {\pi_{t,v}(a_{t,v}\mid s_{t,v})}\right\}
          \left(e_h^{\mathrm H}(\xi_{t,h};\pi)
          -e_h^{\mathrm H}(\xi_{t,h};\pi_t)\right),
        \]
        where $\pi_{t,v}$ is the action distribution of the sampled
        policy $\pi_t$ at step $v$.
      \State Update all candidate weights by
        $\displaystyle
          w_{t+1}(\pi)=
          \frac{w_t(\pi)\exp(-\eta\ell_t(\pi))}
          {\sum_{\pi'\in\Pi}w_t(\pi')\exp(-\eta\ell_t(\pi'))}$.
    \EndFor
    \State Return
      $\widehat\pi\in\operatorname*{arg\,max}_{\pi\in\Pi}w_{T+1}(\pi)$.
  \end{algorithmic}
\end{algorithm}

Write $L_{\mathrm H}(\pi)=\Hell^2(P^{\pi},P^{\pi^\star})$ and
$\overline L_{\mathrm H}(\pi)=\E_{P^\pi}\left[
\sum_{h=1}^H e_h^{\mathrm H}(\xi_h;\pi)\right]$.

\begin{theorem}[Online Hellinger guarantee]\label{thm:online-hellinger}
For a finite class $\Pi$, \cref{alg:online-hellinger}
returns $\widehat\pi\in\Pi$ such that, with probability at least
$1-\delta$,
\begin{equation}\label{eq:online-hellinger-bound}
 L_{\mathrm H}(\widehat\pi)\lesssim
 \min_{\pi\in\Pi}L_{\mathrm H}(\pi)+\frac{\log(|\Pi|/\delta)}T.
\end{equation}
\end{theorem}

\smallskip\noindent\emph{Clipping and prefix truncation.}
For a fixed candidate $\pi$, write $P=P^\pi$, $Q=P^{\pi^\star}$,
and $G_h=\sum_{v=1}^h g_v(\xi_v;\pi)$, with $G_0=0$.
For $c>0$, define
\[
 e_h^{(c)}(\xi_h)=\min\{c,G_h\}-\min\{c,G_{h-1}\},
 \qquad \overline L_c=\E_P\left[\min\{c,G_H\}\right].
\]
The proof uses $c=1$ to compare $\overline L_{\mathrm H}(\pi)$
with $L_{\mathrm H}(\pi)$, and $c=1/4$ to bound prefix truncation.
For any trajectory distribution $U$, the auxiliary distribution
$\Rcal_c[U]$ assigns probabilities
\begin{equation}\label{eq:hellinger-report-laws}
 \begin{aligned}
 \Rcal_c[U](h,\xi_h)&=\frac{e_h^{(c)}(\xi_h)}cU(\xi_h),
 \qquad h\in[H],\\
 \Rcal_c[U](\infty,\tau)
 &=\left(1-\frac{G_H}{c}\right)_+U(\tau).
 \end{aligned}
\end{equation}
Equivalently, conditional on $\tau\sim U$, the outcome is
$(h,\xi_h)$ with probability $e_h^{(c)}(\xi_h)/c$ and
$(\infty,\tau)$ with probability $(1-G_H/c)_+$.
These probabilities sum to one because
$\sum_{h=1}^H e_h^{(c)}/c=\min\{1,G_H/c\}$.
The fixed candidate $\pi$ determines $G_h$ and $e_h^{(c)}$ for
all input distributions $U$, including $U=P^{\pi'}$.

\begin{lemma}[Hellinger identities for auxiliary distributions]\label{lem:hellinger-reporting}
For every candidate $\pi$ and $c>0$,
\begin{equation}\label{eq:hellinger-chain-report}
 \begin{aligned}
 L_{\mathrm H}(\pi)
 &=\E_P\left[\sum_{h=1}^H g_h(\xi_h;\pi)\right],\\
 \Hell^2(\Rcal_c[P],\Rcal_c[Q])
 &=\E_P\left[\sum_{h=1}^H
   \left(1-\frac{G_h}{c}\right)_+g_h(\xi_h;\pi)\right]
 \le\overline L_c.
 \end{aligned}
\end{equation}
\end{lemma}
\begin{proof}
Define the prefix affinities
$A_h=\sum_{\xi_h}\sqrt{P(\xi_h)Q(\xi_h)}$ for $h\in[H]$
and $A_{H+1}=\sum_\tau\sqrt{P(\tau)Q(\tau)}$.
The common initial law gives $A_1=1$. For $h\in[H]$,
\begin{align}
 2(A_h-A_{h+1})
 &=\sum_{\xi_h}\sqrt{P(\xi_h)Q(\xi_h)}
   \left(2-2\sum_a\sqrt{\pi_h(a\mid s_h)
                         \pi_h^\star(a\mid s_h)}\right)
 \tag{Common transition laws}\\
 &=\sum_{\xi_h}\sqrt{P(\xi_h)Q(\xi_h)}
   \Hell^2\!\left(\pi_h(\cdot\mid s_h),
                    \pi_h^\star(\cdot\mid s_h)\right)
 \tag{Definition of $\Hell^2$}\\
 &=\E_P\left[g_h(\xi_h;\pi)\right].
 &&(\text{\cref{alg:online-hellinger}})
 \label{eq:hellinger-affinity-increment}
\end{align}
Prefixes with $P(\xi_h)=0$ contribute zero to both sides.
Consequently,
\begin{align*}
 L_{\mathrm H}(\pi)
 &=2(1-A_{H+1})
 \tag{Definition of $\Hell^2$}\\
 &=2\sum_{h=1}^H(A_h-A_{h+1})
 \tag{$A_1=1$}\\
 &=\E_P\left[\sum_{h=1}^H g_h(\xi_h;\pi)\right].
 \tag{\cref{eq:hellinger-affinity-increment}}
\end{align*}

For the auxiliary distributions, the corresponding affinities
before each stopping decision are
\[
 A_h^{(c)}=\sum_{\xi_h}\left(1-\frac{G_{h-1}}c\right)_+
                \sqrt{P(\xi_h)Q(\xi_h)},\qquad
 A_{H+1}^{(c)}=\sum_\tau\left(1-\frac{G_H}c\right)_+
                \sqrt{P(\tau)Q(\tau)}.
\]
Here $A_1^{(c)}=1$. Since
$(1-G_{h-1}/c)_+=e_h^{(c)}/c+(1-G_h/c)_+$,
\begin{align}
 A_h^{(c)}-A_{h+1}^{(c)}
 &=\sum_{\xi_h}\frac{e_h^{(c)}(\xi_h)}c
             \sqrt{P(\xi_h)Q(\xi_h)}
 \notag\\
 &\quad+\sum_{\xi_h}\left(1-\frac{G_h}c\right)_+
             \sqrt{P(\xi_h)Q(\xi_h)}
   \left(1-\sum_a\sqrt{\pi_h(a\mid s_h)
                         \pi_h^\star(a\mid s_h)}\right)
 \tag{Common transition laws}\\
 &=\begin{aligned}[t]
   &\sum_{\xi_h}\frac{e_h^{(c)}(\xi_h)}c
             \sqrt{P(\xi_h)Q(\xi_h)}\\
   &\quad+\frac12\E_P\left[\left(1-\frac{G_h}c\right)_+
                      g_h(\xi_h;\pi)\right].
   \end{aligned}
 \qquad(\text{Definition of $g_h$})
 \label{eq:hellinger-stopped-affinity}
\end{align}

\begin{align*}
 &\Hell^2(\Rcal_c[P],\Rcal_c[Q])\\
 &\quad=2\left(1-A_{H+1}^{(c)}
       -\sum_{h=1}^H\sum_{\xi_h}\frac{e_h^{(c)}(\xi_h)}c
                    \sqrt{P(\xi_h)Q(\xi_h)}\right)
 \tag{\cref{eq:hellinger-report-laws}}\\
 &\quad=2\sum_{h=1}^H\left(A_h^{(c)}-A_{h+1}^{(c)}
       -\sum_{\xi_h}\frac{e_h^{(c)}(\xi_h)}c
                    \sqrt{P(\xi_h)Q(\xi_h)}\right)
 \tag{$A_1^{(c)}=1$}\\
 &\quad=\E_P\left[\sum_{h=1}^H
       \left(1-\frac{G_h}c\right)_+g_h(\xi_h;\pi)\right]
 \tag{\cref{eq:hellinger-stopped-affinity}}\\
 &\quad\le\E_P\left[\sum_{h=1}^H e_h^{(c)}(\xi_h)\right]
       =\overline L_c.
 \tag{$(1-G_h/c)_+g_h\le e_h^{(c)}$}
\end{align*}
The pointwise inequality holds because $G_h<c$ implies
$e_h^{(c)}=g_h$, whereas $G_h\ge c$ makes its left-hand side zero.
\end{proof}

The definition \cref{eq:truncated-clipped-objective}, with
$e_h^{\mathrm H}$ in place of $e_h$, defines
$\operatorname{Trun}_{\pi'}\left[\overline L_{\mathrm H}(\pi)\right]$.
\begin{lemma}[Hellinger clipping and truncation]\label{lem:hellinger-comparison}
For every $\pi,\pi'\in\Pi$,
\begin{equation}\label{eq:hellinger-comparison}
 \overline L_{\mathrm H}(\pi)\asymp L_{\mathrm H}(\pi),\qquad
 \overline L_{\mathrm H}(\pi)\lesssim
 \operatorname{Trun}_{\pi'}\!\left[\overline L_{\mathrm H}(\pi)\right]
 +\overline L_{\mathrm H}(\pi').
\end{equation}
More precisely,
\begin{equation}\label{eq:hellinger-truncation-remainder}
 \operatorname{Trun}_{\pi'}\left[\overline L_{\mathrm H}(\pi)\right]
 \ge\overline L_{\mathrm H}(\pi)/32-L_{\mathrm H}(\pi')/4.
\end{equation}
Also $\overline L_{\mathrm H}(\pi)\le L_{\mathrm H}(\pi)$ and
$0\le\operatorname{Trun}_{\pi'}[\overline L_{\mathrm H}(\pi)]
\le\overline L_{\mathrm H}(\pi)$.
\end{lemma}
\begin{proof}
Clipping and \cref{eq:hellinger-chain-report} give
$\overline L_{\mathrm H}(\pi)\le L_{\mathrm H}(\pi)$.
The bounds
$0\le\operatorname{Trun}_{\pi'}[\overline L_{\mathrm H}(\pi)]
\le\overline L_{\mathrm H}(\pi)$ follow from
\cref{eq:truncated-clipped-objective}.
For the reverse inequality, let
$D=\{(h,\xi_h):h\in[H]\}$ be the set of prefix outcomes.
Then $\Rcal_1[P](D)=\overline L_{\mathrm H}(\pi)$, and
\begin{align}
 \sqrt{\E_Q\left[\min\{1,G_H\}\right]}
 &=\sqrt{\Rcal_1[Q](D)}
 \tag{\cref{eq:hellinger-report-laws}}\\
 &\le\sqrt{\Rcal_1[P](D)}
       +\Hell(\Rcal_1[P],\Rcal_1[Q])
 \tag{\cref{eq:weighted-hell}, $F=\1_D$}\\
 &\le2\sqrt{\overline L_{\mathrm H}(\pi)}.
 &&(\text{\cref{eq:hellinger-chain-report}})
 \label{eq:hellinger-teacher-clipped}
\end{align}
Splitting each trajectory's contribution with
$1=(1-G_H)_++\min\{1,G_H\}$ gives
\begin{align}
 L_{\mathrm H}(\pi)
 &=\sum_\tau\left(
   \sqrt{\Rcal_1[P](\infty,\tau)}
   -\sqrt{\Rcal_1[Q](\infty,\tau)}\right)^2
 \notag\\
 &\quad+\sum_\tau\min\{1,G_H\}
                 (\sqrt{P(\tau)}-\sqrt{Q(\tau)})^2
 \tag{\cref{eq:hellinger-report-laws}}\\
 &\le\Hell^2(\Rcal_1[P],\Rcal_1[Q])
       +\overline L_{\mathrm H}(\pi)
       +\E_Q\left[\min\{1,G_H\}\right]
 \tag{$(\sqrt p-\sqrt q)^2\le p+q$}\\
 &\le6\overline L_{\mathrm H}(\pi).
 &&(\text{\cref{eq:hellinger-chain-report,eq:hellinger-teacher-clipped}})
 \label{eq:hellinger-clipped-conversion}
\end{align}

For truncation, set $c=1/4$, $R=\Rcal_c[P]$,
$S=\Rcal_c[Q]$, $S'=\Rcal_c[P^{\pi'}]$, and
$z=R(D)=\overline L_c/c$.
The overlap inequality first compares $R$ with $S$, and then
$S$ with $S'$:
\begin{align}
 \frac12\sqrt z
 &\le\sqrt{\sum_{y\in D}\min\{R(y),S(y)\}}
 \tag{\cref{eq:overlap-triangle,eq:hellinger-chain-report}, $c=1/4$}\\
 &\le\sqrt{\sum_{y\in D}\min\{R(y),S'(y)\}}+\Hell(S,S')
 \tag{\cref{eq:overlap-triangle}}\\
 &\le\sqrt{\sum_{y\in D}\min\{R(y),S'(y)\}}
       +\sqrt{L_{\mathrm H}(\pi')}.
 &&(\text{\cref{eq:auxiliary-hellinger-contraction}})
 \label{eq:hellinger-shared-prefix}
\end{align}
Since $e_h^{(c)}\le e_h^{\mathrm H}$ and
$\min\{c,G_H\}\ge c\min\{1,G_H\}$,
\begin{align*}
 &\operatorname{Trun}_{\pi'}\left[\overline L_{\mathrm H}(\pi)\right]\\
 &\quad\ge\sum_{h=1}^H\sum_{\xi_h}
       \min\{P(\xi_h),P^{\pi'}(\xi_h)\}e_h^{(c)}(\xi_h)
 \tag{$e_h^{(c)}\le e_h^{\mathrm H}$}\\
 &\quad=c\sum_{y\in D}\min\{R(y),S'(y)\}
 \tag{\cref{eq:hellinger-report-laws}}\\
 &\quad\ge c\left(\frac z8-L_{\mathrm H}(\pi')\right)
 \tag{\cref{eq:hellinger-shared-prefix}, $(x+y)^2\le2x^2+2y^2$}\\
 &\quad=\frac{\overline L_c}{8}-\frac{L_{\mathrm H}(\pi')}{4}
       \ge\frac{\overline L_{\mathrm H}(\pi)}{32}
          -\frac{L_{\mathrm H}(\pi')}{4}.
 \tag{$cz=\overline L_c$, $\overline L_c\ge c\overline L_{\mathrm H}(\pi)$}
\end{align*}
This proves \cref{eq:hellinger-truncation-remainder}.
Rearranging \cref{eq:hellinger-truncation-remainder} and using
$L_{\mathrm H}(\pi')\le6\overline L_{\mathrm H}(\pi')$ from
\cref{eq:hellinger-clipped-conversion} gives
\[
 \overline L_{\mathrm H}(\pi)
 \le32\operatorname{Trun}_{\pi'}
       \left[\overline L_{\mathrm H}(\pi)\right]
       +48\overline L_{\mathrm H}(\pi'),
\]
which proves the second relation in \cref{eq:hellinger-comparison}.
\end{proof}

\begin{proof}[Proof of \cref{thm:online-hellinger}]
The conditional-mean, cancellation, and second-moment calculations
in \cref{lem:weight-moments} use only the form of the loss
and nonnegative increments satisfying $\sum_{h=1}^H e_h\le1$.
Replacing $e_h$ by $e_h^{\mathrm H}$ therefore gives
$|\ell_t(\pi)|\le1$ and
\cref{eq:online-loss-mean}, \cref{eq:weighted-loss-zero}, and
\cref{eq:loss-moments} with
$\overline L_{\mathrm H}$ in place of $\overline L$.

Applying \cref{eq:hellinger-truncation-remainder} to each direction
of the conditional mean in \cref{eq:online-loss-mean}, and using
$\overline L_{\mathrm H}\le L_{\mathrm H}\le6\overline L_{\mathrm H}$,
gives
\[
 \begin{aligned}
 -\E_t\left[\ell_t(\pi)\right]
 &\ge\frac1{32}\sum_{\pi'\in\Pi}w_t(\pi')
       \overline L_{\mathrm H}(\pi')-\frac54L_{\mathrm H}(\pi),\\
 \E_t\left[\ell_t(\pi)\right]
 &\ge\frac1{32}\overline L_{\mathrm H}(\pi)
       -\frac52\sum_{\pi'\in\Pi}w_t(\pi')
                     \overline L_{\mathrm H}(\pi').
 \end{aligned}
\]
Thus \cref{eq:loss-risk-drift} holds with $a=1/32$, $C=5/2$,
and $L_{\mathrm H},\overline L_{\mathrm H}$ in place of
$L,\overline L$, since $5/4\le2$.
The stepsize $\eta=1/128$ satisfies $a-2\eta=1/64>0$.
The analysis of concentration and selection by final weight in
\cref{app:online-ew}, from \cref{eq:ew-standard} through
\cref{eq:final-clipped-risk}, therefore gives, with probability
at least $1-\delta$,
\[
 L_{\mathrm H}(\widehat\pi)
 \le6\overline L_{\mathrm H}(\widehat\pi)
 \lesssim\min_{\pi\in\Pi}L_{\mathrm H}(\pi)
          +\frac{\log(|\Pi|/\delta)}T,
\]
where the first inequality is \cref{eq:hellinger-clipped-conversion}.
\end{proof}

\begin{corollary}[Reverse KL via online Hellinger learning]\label{cor:online-hellinger-kl}
Under \cref{ass:ratio-cap}, \cref{alg:online-hellinger}
satisfies, with probability at least $1-\delta$,
\begin{equation}\label{eq:online-hellinger-kl}
 L(\widehat\pi)\lesssim B_\infty
 \left(\min_{\pi\in\Pi}\Hell^2(P^\pi,P^{\pi^\star})
 +\frac{\log(|\Pi|/\delta)}T\right).
\end{equation}
\end{corollary}
\begin{proof}
\cref{thm:online-hellinger} and
\cref{lem:ratio-conversion} give \cref{eq:online-hellinger-kl}.
Moreover, $\min_{\pi\in\Pi}L_{\mathrm H}(\pi)\le L^\star$ by
\cref{eq:hellinger-kl}, recovering the reverse-KL guarantee with
approximation term $B_\infty L^\star$.
\end{proof}

\Needspace{22\baselineskip}
\subsection{Estimation and approximation lower bounds}\label{app:optimality-lower-bounds}
\begin{optimalitystatement}
\textbf{Estimation.}
For every integer $d\ge1$, $T\ge1$, $B\ge1$, and
$\delta\in(0,1/8]$, every online or offline full-feedback learner
using $T$ episodes has a realizable instance with $H=1$,
$|\mathcal A|=2$, and $|\Pi|=2^d$ satisfying \cref{ass:relative-second-moment}
on which, with probability greater than $\delta$,
\begin{equation}\label{eq:realizable-lower}
 L(\widehat\pi)=\Omega\!\left(
 B\min\!\left\{1,\frac{\log(|\Pi|/\delta)}T\right\}\right).
\end{equation}
Under \cref{ass:ratio-cap}, the same lower bound holds with
$B_\infty>1$ replacing $B$, including for offline action feedback.

\smallskip\noindent\textbf{Approximation at optimal estimation.}
Fix $\kappa>0$. There is an integer $k=k(\kappa)\ge8$ such that,
for $B\ge2\log k$ and $T\ge k$, any learner returning
$\widehat\pi\in\Pi$ whose guarantee at confidence $3/4$ has
$C_{\rm est}\le\kappa B$ on all $H=2$ instances satisfying
\cref{ass:relative-second-moment} with $|\mathcal A|=|\Pi|=k$ must have
$\Capx=\Omega(B)$.
This holds for both online and offline full feedback.
Under \cref{ass:ratio-cap}, the same conclusion holds with
$B_\infty$ replacing $B$, including for offline action feedback.
\end{optimalitystatement}

\subsubsection{Estimation lower bound}\label[appendix]{app:estimation}
\label{app:estimation-lower-bounds}\label{app:confidence}
\begin{proof}[Proof of the estimation lower bound]
\emph{MDP and policy class.}
The one-step MDP has
\[
 \mathcal A=\{a_1,a_2\},\qquad
 \mathcal S_1=\{s_{1,0},s_{1,1},\ldots,s_{1,d}\},\qquad
 \Pi=\{\pi_f:f\in\{0,1\}^d\}.
\]
For $\rho\in(0,1]$, the initial law is
$\mu(s_{1,x})=\rho/d$ for $x\in[d]$ and $\mu(s_{1,0})=1-\rho$.
The episode ends after the first action.
For $\beta\ge1$, define distributions on $(a_1,a_2)$ by
\[
 R_0(\beta)=\left(\frac{e^\beta}{1+e^\beta},
                         \frac1{1+e^\beta}\right),\qquad
 R_1(\beta)=\left(\frac1{1+e^\beta},
                         \frac{e^\beta}{1+e^\beta}\right).
\]
For $f\in\{0,1\}^d$ and $x\in[d]$,
\[
 \pi_{f,1}(\cdot\mid s_{1,x})=R_{f_x}(\beta),\qquad
 \pi_{f,1}(\cdot\mid s_{1,0})=(1/2,1/2).
\]
The teacher is $\pi^\star_\theta=\pi_\theta$ for $\theta\in\{0,1\}^d$, so $L^\star=0$.

\emph{Boundedness.}
For $x\in[d]$ and each feasible $\tau=(s_1,a_1)$,
\begin{align*}
 k_1(s_{1,x};\pi_f)
 &=\KL(R_{f_x}(\beta)\|R_{\theta_x}(\beta))
 =\beta\tanh(\beta/2)\1\{f_x\ne\theta_x\}\le\beta,
 \tag{Bernoulli KL formula}\\
 \frac{P^{\pi_f}(\tau)}{P^{\pi^\star_\theta}(\tau)}
 &=\frac{\pi_{f,1}(a_1\mid s_1)}
         {\pi^\star_{\theta,1}(a_1\mid s_1)}
 \in\{1,e^{-\beta},e^\beta\}.
 \tag{Common initial law}
\end{align*}
Thus $\beta=B$ verifies \cref{ass:relative-second-moment} by \cref{lem:second-moment-sufficient}, and $\beta=B_\infty$
verifies \cref{ass:ratio-cap}, for either initial law used below.

\emph{Risk at an unvisited state.}
Write $L_\theta(\pi)=\KL(P^\pi\|P^{\pi^\star_\theta})$.
Let $\theta$ be uniform on $\{0,1\}^d$, and let $M_x$ count
training episodes starting at $s_{1,x}$.
For any binary distribution $v$ and $U_2=(1/2,1/2)$,
\begin{equation}\label{eq:binary-uncertainty}
 \frac12\sum_{z=0}^1\KL(v\|R_z(\beta))
 =\KL(v\|U_2)+b(\beta)\ge b(\beta),
 \qquad b(\beta):=\log\cosh(\beta/2)\asymp\beta.
\end{equation}
Let $v_x=\widehat\pi_1(\cdot\mid s_{1,x})$.
Conditional on $M_x=0$, the training observations and $v_x$
are independent of $\theta_x$ for either trajectory source.
\cref{eq:binary-uncertainty} gives
\[
 \Prob\!\left\{\KL(v_x\|R_{\theta_x}(\beta))\ge b(\beta)
       \mid M_x=0\right\}\ge\frac12.
\]

\emph{Class-size term.}
Take $\rho=\min\{1,d/(8T)\}$ and set
\[
 W=\frac1d\sum_{x=1}^d
 \1\{M_x=0,\ \KL(v_x\|R_{\theta_x}(\beta))\ge b(\beta)\}.
\]
Then
\begin{align*}
 \E\left[W\right]
 &\ge\frac1{2d}\sum_{x=1}^d\Prob\{M_x=0\}
 =\frac12\left(1-\frac\rho d\right)^T
 \tag{Independent initial-state draws}\\
 &\ge\frac12\left(1-\frac{T\rho}d\right)\ge\frac7{16},
 \tag{Bernoulli's inequality}\\
 L_\theta(\widehat\pi)
 &=\sum_{s\in\mathcal S_1}\mu(s)
   \KL(\widehat\pi_1(\cdot\mid s)\|\pi^\star_{\theta,1}(\cdot\mid s))
 \tag{$H=1$}\\
 &\ge\frac\rho d\sum_{x=1}^d
       \KL(v_x\|R_{\theta_x}(\beta))
 \ge\rho b(\beta)W,
 \tag{Definition of $W$}\\
 \Prob\!\left\{L_\theta(\widehat\pi)>\frac{\rho b(\beta)}8\right\}
 &\ge\Prob\{W>1/8\}
 \ge\frac{\E\left[W\right]-1/8}{1-1/8}
 \ge\frac5{14}>\frac14.
 \tag{$0\le W\le1$}
\end{align*}
Here $\rho b(\beta)\asymp\beta\min\{1,\log|\Pi|/T\}$.
Averaging over the finite teacher prior gives a fixed teacher
with this failure probability.

\emph{Confidence term.}
Keep the same policy class and take
$\rho=\min\{1/2,\log(1/(2\delta))/(4T)\}$.
Replace the initial law by
\[
 \mu(s_{1,1})=\rho,\qquad
 \begin{cases}
 \mu(s_{1,0})=1-\rho,&d=1,\\
 \mu(s_{1,0})=0,\quad
 \mu(s_{1,x})=(1-\rho)/(d-1)\quad(2\le x\le d),&d\ge2.
 \end{cases}
\]
Only $\theta_1$ is unknown: it is uniform on $\{0,1\}$, and
$\theta_x=0$ for $x\ge2$.
Thus observations away from $s_{1,1}$ reveal no information about
$\theta_1$, and
\begin{align*}
 \Prob\{L_\theta(\widehat\pi)\ge\rho b(\beta)\}
 &\ge\tfrac12\Prob\{M_1=0\}
 =\tfrac12(1-\rho)^T
 \tag{\cref{eq:binary-uncertainty}}\\
 &\ge\tfrac12e^{-2T\rho}
 \ge\tfrac12\sqrt{2\delta}>\delta.
 \tag{$\log(1-\rho)\ge-2\rho$, $\delta\le1/8$}
\end{align*}
Here $\rho b(\beta)\asymp\beta\min\{1,\log(1/\delta)/T\}$,
and prior averaging again fixes a teacher.
Choosing the larger of the two risk thresholds yields
\[
 \beta\max\!\left\{\min\!\left\{1,\frac{\log|\Pi|}T\right\},
                      \min\!\left\{1,\frac{\log(1/\delta)}T\right\}\right\}
 \asymp\beta\min\!\left\{1,\frac{\log(|\Pi|/\delta)}T\right\}.
\]
Both arguments allow full feedback from either trajectory source.
Restricting feedback gives the offline action-feedback bound.
\end{proof}

\subsubsection{Approximation lower bound}\label{app:approximation-lower-bounds}
\label[appendix]{app:proper}\label{app:proper-ratio}
\begin{proof}[Proof of the approximation lower bound]
We use a construction parameter $B\ge2\log k$.
\emph{MDP and policy class.}
Let $k\ge8$ be chosen below in terms of $\kappa$, and set
$\rho=(k-1)/(8T)$. The two-step MDP has
\[
 \mathcal A=\{a_1,\ldots,a_k\},\qquad
 \mathcal S_1=\{s_{1,0},s_{1,1}\},\qquad
 \mathcal S_2=\{s_{2,0},s_{2,1},\ldots,s_{2,k}\},
\]
\[
 \mu(s_{1,1})=\rho,\qquad \mu(s_{1,0})=1-\rho,
\]
\[
 \mathsf P_1(s_{2,j}\mid s_{1,1},a_j)=1,\qquad
 \mathsf P_1(s_{2,0}\mid s_{1,0},a_j)=1\quad(j\in[k]).
\]
The episode ends after the second action.
The class is $\Pi=\{\pi_i:i\in[k]\}$, and the teacher is
indexed by $\theta\in[k]$.
Their first-step distributions at $s_{1,1}$ are
\[
 \pi_{i,1}(a_j\mid s_{1,1})
 =\begin{cases}3/4,&j=i,\\1/[4(k-1)],&j\ne i,\end{cases}
 \qquad
 \pi^\star_{\theta,1}(a_j\mid s_{1,1})=1/k.
\]
For $z\in\{0,1\}$, define distributions on $\mathcal A$ by
\[
 R_z(a_1)=\frac{\exp((B/2)\1\{z=0\})}{1+e^{B/2}},\qquad
 R_z(a_j)=\frac{\exp((B/2)\1\{z=1\})}
                  {(1+e^{B/2})(k-1)}\quad(2\le j\le k).
\]
At the second layer,
\[
 \pi_{i,2}(\cdot\mid s_{2,j})=R_{\1\{j=i\}},\qquad
 \pi^\star_{\theta,2}(\cdot\mid s_{2,j})=R_{\1\{j=\theta\}}
 \quad(j\in[k]).
\]
Every policy is uniform at $s_{1,0}$ and $s_{2,0}$.
The learner may know the initial law and transitions and explore
adaptively, but cannot reset within an episode or query unvisited
states. Its output belongs to $\Pi$.

\emph{Boundedness and trajectory risks.}
Write $L_\theta(\pi)=\KL(P^\pi\|P^{\pi^\star_\theta})$ and
$d_k=\KL(\Ber(3/4)\|\Ber(1/k))\le\log k$.
For $B\ge2\log k$,
\begin{align*}
 \KL(R_0\|R_1)=\KL(R_1\|R_0)
 &=\frac{B}{2}\tanh(B/4)\in[B/4,B/2],
 \tag{Bernoulli KL formula, $k\ge8$}\\
 k_1(s_{1,1};\pi_i)&=d_k,\qquad
 k_2(s_{2,j};\pi_i)
 =\KL(R_0\|R_1)\1\{\1\{j=i\}\ne\1\{j=\theta\}\}.
 \tag{The specified action distributions}
\end{align*}
All action probabilities are positive, and every feasible trajectory
satisfies
\begin{align*}
 \sum_{h=1}^2 k_h(s_h;\pi_i)
 &\le\log k+B/2\le B,\\
 \log\frac{P^{\pi_i}(\tau)}{P^{\pi^\star_\theta}(\tau)}
 &=\sum_{h=1}^2
   \log\frac{\pi_{i,h}(a_h\mid s_h)}{\pi^\star_{\theta,h}(a_h\mid s_h)}
 \le\log k+B/2\le B.
 \tag{First-step ratio $\le k$, second-step ratio $\le e^{B/2}$}
\end{align*}
\cref{lem:second-moment-sufficient} verifies
\cref{ass:relative-second-moment} from the first bound.
The second verifies \cref{ass:ratio-cap} with $B_\infty=B$.
If $i\ne\theta$, only branches $j=i$ and $j=\theta$ have
nonzero second-step KL. The KL chain rule gives
\begin{equation}\label{eq:proper-risk}
 \begin{aligned}
 L_\theta(\pi_i)
 &=\rho d_k+\rho\left(\frac34+\frac1{4(k-1)}\right)
       \KL(R_0\|R_1)\1\{i\ne\theta\},\\
 L^\star&=L_\theta(\pi_\theta)=\rho d_k.
 \end{aligned}
\end{equation}

\emph{Identification error.}
Let $\theta$ be uniform on $[k]$.
The teacher's first-step distribution is independent of $\theta$.
A trajectory taking $a_j$ at $s_{1,1}$ reaches $s_{2,j}$,
where full feedback reveals exactly $\1\{j=\theta\}$.
One equality test also simulates an offline teacher trajectory:
its first action is uniform, and its second follows
$R_{\1\{j=\theta\}}$.
Let $M=\sum_{t=1}^T\1\{s_{t,1}=s_{1,1}\}$ and write
$\widehat\pi=\pi_{\widehat i}$.
Conditional on the initial-state sequence, $M$ is fixed and
$\theta$ remains uniform on $[k]$. Therefore,
\begin{align*}
 \Prob\{\widehat i\ne\theta\}
 &\ge\E\left[1-\frac{M+1}{k}\right]
 \tag{\cref{lem:equality-tests}, conditionally then averaged}\\
 &=\frac{7(k-1)}{8k}\ge\frac34,
 \tag{$\E\left[M\right]=T\rho=(k-1)/8$, $k\ge8$}\\
 L_\theta(\widehat\pi)-L^\star
 &\ge\frac{3\rho B}{16}\quad\text{on }\{\widehat i\ne\theta\}.
 \tag{\cref{eq:proper-risk}, $\KL(R_0\|R_1)\ge B/4$}
\end{align*}
Prior averaging fixes a teacher for which this event has
probability at least $3/4$.

\emph{Approximation coefficient.}
The event above intersects the success event of any guarantee
at confidence $3/4$, so
\begin{align}
 \frac{3\rho B}{16}
 &\le(\Capx-1)L^\star+C_{\rm est}\frac{\log(4k)}T
 \tag{Guarantee at confidence $3/4$}\\
 \frac{3B}{16}
 &\le(\Capx-1)d_k+\frac{8C_{\rm est}\log(4k)}{k-1}.
 &&(\text{$L^\star=\rho d_k$, $\rho=(k-1)/(8T)$})
 \label{eq:proper-tradeoff}
\end{align}
Choose the fixed integer $k=k(\kappa)\ge8$ so that
$8\kappa\log(4k)/(k-1)\le3/32$. Then
\begin{align*}
 \frac{8C_{\rm est}\log(4k)}{k-1}
 &\le\frac{8\kappa B\log(4k)}{k-1}\le\frac{3B}{32},
 \tag{$C_{\rm est}\le\kappa B$}\\
 (\Capx-1)d_k&\ge\frac{3B}{32},
 \tag{\cref{eq:proper-tradeoff}}\\
 \Capx&\ge1+\frac{3B}{32\log k}=\Omega(B).
 \tag{$d_k\le\log k$, fixed $k=k(\kappa)$}
\end{align*}
This proves the relative second-moment result.
Replacing $B$ by $B_\infty$ gives the ratio-bound result.
Restricting the offline observations to action feedback preserves
the lower bound.
\end{proof}

\subsection{The additional cost of online action feedback}\label{app:online-token}
\begin{actionlowerstatement}
For $T\ge1$, $B_\infty>1$, and $|\Pi|,|\mathcal A|\ge2$, let
$m=\min\{|\mathcal A|,|\Pi|,e^{B_\infty}\}$.
Every online action-feedback learner with arbitrary causal exploration
and output in $\Pi$ has a realizable $H=1$ instance with a full-support
candidate class satisfying \cref{ass:ratio-cap} such that,
with probability at least $1/4$,
\begin{equation}\label{eq:online-token-lower-precise}
 L(\widehat\pi)=\Omega\!\left(
 B_\infty\min\!\left\{1,\frac{m\log|\Pi|}{T\log(em)}\right\}\right).
\end{equation}
\end{actionlowerstatement}

\begin{proof}[Proof of \cref{thm:online-token-lower}]
\emph{MDP and policy class.}
Write $A=|\mathcal A|$, $N=|\Pi|$, and set
\[
 m=\min\{A,N,e^{B_\infty}\},\qquad
 k=\lfloor m\rfloor,\qquad d=\lfloor\log N/\log k\rfloor,
 \qquad \rho=\min\!\left\{1,\frac{(k-1)d}{8T}\right\}.
\]
The one-step MDP has
\[
 \mathcal A=\{a_1,\ldots,a_A\},\qquad
 \mathcal S_1=\{s_{1,0},s_{1,1},\ldots,s_{1,d},s_{1,d+1}\},
\]
\[
 \mu(s_{1,x})=\rho/d\quad(x\in[d]),\qquad
 \mu(s_{1,0})=1-\rho,\qquad \mu(s_{1,d+1})=0.
\]
The episode ends after the first action. The state $s_{1,d+1}$
will distinguish additional candidates if $k^d<N$.
For $f\in[k]^d$ and $x\in[d]$, define
\begin{equation}\label{eq:online-token-root}
 \pi_{f,1}(a_j\mid s_{1,x})=
 \begin{cases}
 \displaystyle\frac{A+k}{2A}\,
 \frac{\exp(B_\infty\1\{j=f_x\})}{e^{B_\infty}+k-1},
 &1\le j\le k,\\[6pt]
 \displaystyle\frac1{2A},&k<j\le A.
 \end{cases}
\end{equation}
The last $A-k$ actions have total probability $(A-k)/(2A)$;
the first $k$ have the remaining probability $(A+k)/(2A)\ge1/2$.
At $s_{1,0}$ and $s_{1,d+1}$, every $\pi_f$ is uniform.
The class contains all $k^d$ policies $\pi_f$.
The remaining $N-k^d$ candidates agree with $\pi_{(1,\ldots,1)}$
except at $s_{1,d+1}$, where they use distinct, nonuniform,
strictly positive action distributions.
The teacher is $\pi^\star_\theta=\pi_\theta$, $\theta\in[k]^d$,
so $L^\star=0$.

\emph{Boundedness and trajectory risks.}
Write $L_\theta(\pi)=\KL(P^\pi\|P^{\pi^\star_\theta})$.
All action probabilities are positive. On a trajectory starting
at $s_{1,x}$ with action $a_j$,
\[
 \frac{P^{\pi_f}(\tau)}{P^{\pi^\star_\theta}(\tau)}
 =\begin{cases}
 \exp\!\left(B_\infty[\1\{j=f_x\}-\1\{j=\theta_x\}]\right),
 &x\in[d],\ j\le k,\\
 1,&\text{otherwise}
 \end{cases}
 \le e^{B_\infty}.
\]
This also holds for the additional candidates, whose trajectory
law is $P^{\pi_{(1,\ldots,1)}}$.
For $f_x\ne\theta_x$,
\begin{align*}
 k_1(s_{1,x};\pi_f)
 &=B_\infty\bigl(\pi_{f,1}(a_{f_x}\mid s_{1,x})
                  -\pi_{f,1}(a_{\theta_x}\mid s_{1,x})\bigr)
 \tag{Only $a_{f_x}$ and $a_{\theta_x}$ have nonunit ratios}\\
 &=\frac{A+k}{2A}\,
   \frac{B_\infty(e^{B_\infty}-1)}{e^{B_\infty}+k-1}
 \asymp B_\infty.
 \tag{$k\le e^{B_\infty}$, $B_\infty>1$, $1/2\le(A+k)/(2A)\le1$}
\end{align*}
Consequently,
\begin{equation}\label{eq:online-token-risk}
 L_\theta(\pi_f)
 =\frac{\rho(A+k)}{2Ad}\,
   \frac{B_\infty(e^{B_\infty}-1)}{e^{B_\infty}+k-1}
   \sum_{x=1}^d\1\{f_x\ne\theta_x\}
 \asymp\frac{\rho B_\infty}{d}
   \sum_{x=1}^d\1\{f_x\ne\theta_x\}.
\end{equation}

\emph{Identification error and risk.}
Let $\theta$ be uniform on $[k]^d$.
By \cref{eq:online-token-root}, querying $a_j$ at $s_{1,x}$
reveals exactly $\1\{j=\theta_x\}$ for $j\le k$, and reveals no
information about $\theta_x$ for $j>k$.
Feedback at $s_{1,0}$ is independent of $\theta$.
Let $M_x=\sum_{t=1}^T\1\{s_{t,1}=s_{1,x}\}$ and write the
output as $\pi_{\widehat f}$, assigning label $(1,\ldots,1)$
to the additional candidates.
Conditional on the initial-state sequence and all other labels,
$M_x$ is fixed and $\theta_x$ remains uniform on $[k]$.
\cref{lem:equality-tests}, followed by averaging, gives
\begin{align*}
 \Prob\{\widehat f_x\ne\theta_x\}
 &\ge\E\left[1-\frac{M_x+1}{k}\right]
 \tag{\cref{lem:equality-tests}}\\
 &\ge1-\frac{(k-1)/8+1}{k}
 =\frac{7(k-1)}{8k}\ge\frac7{16}.
 \tag{$\E\left[M_x\right]=T\rho/d\le(k-1)/8$, $k\ge2$}
\end{align*}
For $W=d^{-1}\sum_{x=1}^d\1\{\widehat f_x\ne\theta_x\}$,
\begin{align*}
 \E\left[W\right]&\ge7/16,
 \qquad
 \Prob\{W>1/8\}
 \ge\frac{\E\left[W\right]-1/8}{1-1/8}
 \ge\frac5{14},
 \tag{$0\le W\le1$}\\
 L_\theta(\widehat\pi)
 &\gtrsim\rho B_\infty
 \quad\text{on }\{W>1/8\},
 \tag{\cref{eq:online-token-risk}}\\
 \rho B_\infty
 &\asymp B_\infty\min\!\left\{1,\frac{kd}{T}\right\}
 \tag{$k\ge2$ and the definition of $\rho$}\\
 &\asymp B_\infty\min\!\left\{1,
       \frac{m\log N}{T\log(em)}\right\}.
 \tag{$k\asymp m$, $d\asymp\log N/\log k$, $\log k\asymp\log(em)$}
\end{align*}
Averaging over the finite teacher prior gives a fixed teacher
with failure probability at least $5/14>1/4$.
\end{proof}

%% file: appendix_offline.tex
\section{Offline learning}\label[appendix]{app:offline}
\subsection{Full feedback}\label{app:offline-full-proof}
\begin{proof}[Proof of \cref{prop:offline-unclipped}]
If $\pi^\star\in\Pi$, both empirical objectives have minimum zero.
Nonnegativity of the local KL losses gives, for every $\pi\in\Pi$,
\[
 \sum_{t=1}^T\sum_{h=1}^H k_h(s_{t,h}^\star;\pi)=0
 \quad\Longleftrightarrow\quad
 \sum_{t=1}^T\min\!\left\{1,
     \sum_{h=1}^H k_h(s_{t,h}^\star;\pi)\right\}=0.
\]
Thus \cref{eq:offline-unclipped-estimator} and
\cref{eq:offline-full-estimator} have the same minimizers, and
\cref{thm:offline-full} with $L^\star=0$ proves the first claim.
The lower bound \cref{eq:realizable-lower} shows that this estimation rate is
minimax optimal.

\paragraph{MDP and policy class.}
For the misspecified instance, fix $B\ge32$ and set
\[
 \epsilon=\frac1{B},\qquad
 p=\frac1{16\log B},\qquad q=\frac p{B}.
\]
The $H=2$ MDP has action set $\mathcal A=\{a_1,a_2\}$ and
state layers $\mathcal S_1=\{s_{1,1},s_{1,2}\}$ and
$\mathcal S_2=\{s_{2,0},s_{2,1},s_{2,2}\}$.
The initial law is $\mu(s_{1,1})=\mu(s_{1,2})=1/2$, and
\[
 \mathsf P_1(s_{2,j}\mid s_{1,j},a_1)=1,\qquad
 \mathsf P_1(s_{2,0}\mid s_{1,j},a_2)=1\quad(j\in\{1,2\}).
\]
The episode ends after the second action. The table gives the
probability of $a_1$ at each state, with its complement assigned to $a_2$.
The candidate class is $\Pi=\{\pi_{\rm good},\pi_{\rm bad}\}$.
\[
\begin{array}{c|cc|ccc}
 &s_{1,1}&s_{1,2}&s_{2,0}&s_{2,1}&s_{2,2}\\ \hline
 \pi^\star&\epsilon&q&0&e^{-B/2}&e^{-B/2}\\
 \pi_{\rm good}&0&q&0&1&e^{-B/2}\\
 \pi_{\rm bad}&\epsilon&p&0&e^{-B/2}&1
\end{array}
\]

\paragraph{Boundedness.}
The only nonzero local KL losses are
\[
\begin{aligned}
 k_1(s_{1,1};\pi_{\rm good})
 &=\ell:=-\log(1-\epsilon),&
 k_2(s_{2,1};\pi_{\rm good})&=B/2,\\
 k_1(s_{1,2};\pi_{\rm bad})
 &=r:=p\log\frac pq+(1-p)\log\frac{1-p}{1-q},&
 k_2(s_{2,2};\pi_{\rm bad})&=B/2.
\end{aligned}
\]
Since $0<q<p<1$,
\begin{align*}
 \epsilon\le\ell&\le2\epsilon,
 &0\le r&\le p\log(p/q)=\frac1{16},
 \tag{$\epsilon\le1/2$, $p/q=B$}\\
 \sum_{h=1}^2 k_h(s_h;\pi)
 &\le\frac{B}{2}+\max\{\ell,r\}\le B
 &&\text{for every $\pi\in\Pi$ and $\tau\in\mathcal T_{\rm env}$.}
\end{align*}
This verifies the pathwise condition of
\cref{lem:second-moment-sufficient}, including trajectories that
a candidate never generates, and hence
\cref{ass:relative-second-moment}.

\paragraph{Population objectives and empirical selection.}
Under the teacher, $s_{2,1}$ and $s_{2,2}$ have probabilities
$\epsilon/2$ and $q/2$. Under $\pi_{\rm good}$, $s_{2,1}$ is never
visited, whereas $\pi_{\rm bad}$ visits $s_{2,2}$ with probability $p/2$.
Consequently,
\[
\begin{array}{c|cc}
 &\displaystyle\E_{P^{\pi^\star}}\left[
       \sum_{h=1}^2 k_h(s_h;\pi)\right]&L(\pi)\\[3pt] \hline
 \pi_{\rm good}&(\ell+\epsilon B/2)/2&\ell/2\\[2pt]
 \pi_{\rm bad}&(r+qB/2)/2&(r+pB/2)/2
\end{array}
\]
The untruncated objective ranks the candidates in the wrong order,
\begin{align*}
 \E_{P^{\pi^\star}}\left[
       \sum_{h=1}^2 k_h(s_h;\pi_{\rm bad})\right]
 &\le\frac1{32}+\frac p4\le\frac1{16},
 \tag{$r\le1/16$, $q=p/B$}\\
 \E_{P^{\pi^\star}}\left[
       \sum_{h=1}^2 k_h(s_h;\pi_{\rm good})\right]
 &\ge\frac{\epsilon B}4=\frac14,
 \tag{$\ell\ge0$, $\epsilon=1/B$}\\
 \frac{L(\pi_{\rm bad})}{L(\pi_{\rm good})}
 &=\frac{r+pB/2}{\ell}
 \ge\frac{pB}{4\epsilon}
 =\frac{B^2}{64\log B}>1.
 \tag{$r\ge0$, $\ell\le2\epsilon$, $B\ge32$}
\end{align*}
In particular, $L^\star=L(\pi_{\rm good})>0$.
For each $\pi\in\Pi$, the strong law of large numbers gives,
with probability one,
\[
 \frac1T\sum_{t=1}^T\sum_{h=1}^2 k_h(s_{t,h}^\star;\pi)
 \longrightarrow
 \E_{P^{\pi^\star}}\!\left[\sum_{h=1}^2 k_h(s_h;\pi)\right].
\]
The population objective of $\pi_{\rm bad}$ is smaller by at
least $3/16$, so
\[
 \Prob\{\widehat\pi_{\rm raw}=\pi_{\rm bad}\}\longrightarrow1.
\]
Consequently, any guarantee with an estimation term vanishing as
$T\to\infty$ requires
\[
 \Capx\ge\frac{L(\pi_{\rm bad})}{L^\star}
 \ge\frac{B^2}{64\log B}
 =\Omega\!\left(\frac{B^2}{\log(eB)}\right).
\]
\end{proof}

\begin{proof}[Proof of \cref{thm:offline-full}]
Each teacher trajectory contributes
$\sum_{h=1}^H e_h(\xi_{t,h}^\star;\pi)\in[0,1]$ to
\cref{eq:offline-full-estimator}, and the teacher trajectories
are independent. Let
$\pi^\dagger\in\arg\min_{\pi\in\Pi}L(\pi)$.
With probability at least $1-\delta$,
\begin{align}
 \overline L(\widehat\pi)
 &\lesssim\E_{P^{\pi^\star}}\left[
                \sum_{h=1}^H e_h(\xi_h;\widehat\pi)\right]
 \tag{\cref{eq:visibility-main}}\\
 &\lesssim\min_{\pi\in\Pi}\E_{P^{\pi^\star}}\left[
                \sum_{h=1}^H e_h(\xi_h;\pi)\right]
          +\frac{\log(|\Pi|/\delta)}T
 \tag{\cref{lem:bounded-score-erm}}\\
 &\le\E_{P^{\pi^\star}}\left[
                \sum_{h=1}^H e_h(\xi_h;\pi^\dagger)\right]
          +\frac{\log(|\Pi|/\delta)}T
 \tag{$\pi^\dagger\in\Pi$}\\
 &\lesssim\overline L(\pi^\dagger)+\frac{\log(|\Pi|/\delta)}T
 \tag{\cref{eq:visibility-main}}\\
 &\le L^\star+\frac{\log(|\Pi|/\delta)}T.
 &&(\text{$\overline L(\pi^\dagger)\le L(\pi^\dagger)=L^\star$})
 \label{eq:offline-full-clipped-risk}
\end{align}
The expectations under $P^{\pi^\star}$ evaluate each displayed
candidate on an independent teacher trajectory.
Under \cref{ass:relative-second-moment},
\begin{align*}
 L(\widehat\pi)
 &\le(1+B)\overline L(\widehat\pi)
 \tag{\cref{lem:second-moment-clipping}}\\
 &\lesssim B\left[L^\star+\frac{\log(|\Pi|/\delta)}T\right].
 \tag{\cref{eq:offline-full-clipped-risk}}
\end{align*}
\end{proof}

\subsection{Action feedback}\label{app:offline-action-proof}

\begin{proposition}[Insufficiency of the relative second-moment bound]
\label{prop:offline-action-B-lower}
For every $B\ge1$ and integer $T\ge1$, there exist an $H=1$ MDP
with one state and three actions and a class $\Pi$ of two policies
such that every possibly randomized offline action-feedback learner
returning $\widehat\pi\in\Pi$ has a teacher $\pi^\star\in\Pi$
for which \cref{ass:relative-second-moment} holds and
\[
 L^\star=0,\qquad
 \Prob\!\left\{L(\widehat\pi)\ge B/2\right\}\ge\frac14.
\]
\end{proposition}
\begin{proof}
Fix $B\ge1$ and $T\ge1$.
\paragraph{MDP and policy class.}
The state and action sets are $\mathcal S=\{s_{1,1}\}$ and
$\mathcal A=\{a_1,a_2,a_3\}$, with $\mu(s_{1,1})=1$.
The episode ends after the first action. Let
\[
 \epsilon=\frac1{4T},\qquad r=\epsilon e^{-B/\epsilon},
 \qquad \Pi=\{\pi_0,\pi_1\},
\]
where the action probabilities at $s_{1,1}$ are
\[
\begin{array}{c|ccc}
 &a_1&a_2&a_3\\ \hline
 \pi_0&1-\epsilon-r&\epsilon&r\\
 \pi_1&1-\epsilon-r&r&\epsilon
\end{array}
\]
The teacher is $\pi^\star=\pi_\theta$ for an unknown
$\theta\in\{0,1\}$.

\paragraph{Boundedness and candidate risks.}
All action probabilities are positive, and for either teacher,
\begin{align}
 k_1(s_{1,1};\pi_{1-\theta})
 &=\KL(P^{\pi_{1-\theta}}\|P^{\pi_\theta})
 =\epsilon\log\frac\epsilon r+r\log\frac r\epsilon
 \tag{The two policies agree on $a_1$}\\
 &=(\epsilon-r)\frac B\epsilon
 =B\bigl(1-e^{-B/\epsilon}\bigr)\in[B/2,B].
 &&(\text{$B/\epsilon=4BT\ge4$})
 \label{eq:offline-action-B-risk}
\end{align}
Since $k_1(s_{1,1};\pi_\theta)=0$, both candidates satisfy $K_\pi\le B$ pointwise.
\cref{lem:second-moment-sufficient} verifies \cref{ass:relative-second-moment},
and $L^\star=L(\pi_\theta)=0$.

\paragraph{Indistinguishable observations.}
Write $\Prob_\theta$ for the law of the training data and the
learner's output when $\pi^\star=\pi_\theta$.
For $E=\{a_{t,1}^\star=a_1\text{ for all }t\in[T]\}$,
\begin{align*}
 \Prob_\theta(E)
 &=(1-\epsilon-r)^T
 \tag{Independent teacher trajectories}\\
 &\ge1-T(\epsilon+r)\ge1-2T\epsilon=\frac12.
 \tag{Bernoulli's inequality, $r\le\epsilon$}
\end{align*}
On $E$, both teachers produce the same $T$ observations
$(s_{1,1},a_1,1-\epsilon-r)$, including the reported action
probabilities. Thus the conditional output distributions coincide,
and $\widehat\pi\in\{\pi_0,\pi_1\}$ implies
\[
 \Prob_0\!\left(\widehat\pi\ne\pi_0\mid E\right)
 +\Prob_1\!\left(\widehat\pi\ne\pi_1\mid E\right)=1.
\]
Consequently,
\begin{align*}
 \max_{\theta\in\{0,1\}}\Prob_\theta\!\left\{
 \KL(P^{\widehat\pi}\|P^{\pi_\theta})\ge B/2\right\}
 &=\max_{\theta\in\{0,1\}}
   \Prob_\theta\!\left\{\widehat\pi\ne\pi_\theta\right\}
 \tag{\cref{eq:offline-action-B-risk}}\\
 &\ge\frac12\sum_{\theta=0}^1
   \Prob_\theta\!\left(\{\widehat\pi\ne\pi_\theta\}\cap E\right)
 \tag{Maximum bounds the average}\\
 &=\frac{(1-\epsilon-r)^T}{2}\ge\frac14.
 \tag{Identical observations on $E$}
\end{align*}
\end{proof}

\begin{proof}[Proof of \cref{thm:offline-token}]
The estimator in \citet[Appendix C, Eq.~(11)]{rohatgi2025}
maximizes the empirical log-likelihood of
$(P^\pi+P^{\pi^\star})/2$ over $\pi\in\Pi$.
Removing terms independent of $\pi$ gives the equivalent objective
\begin{equation}\label{eq:offline-token-alg}
 \widehat\pi\in\operatorname*{arg\,min}_{\pi\in\Pi}
 -\sum_{t=1}^T\log\!\left(
 1+\frac{P^\pi(\tau_t^\star)}{P^{\pi^\star}(\tau_t^\star)}\right).
\end{equation}
For every observed teacher trajectory, the common initial-state and
transition probabilities cancel, so
\begin{equation}\label{eq:offline-ratio}
 \frac{P^\pi(\tau_t^\star)}{P^{\pi^\star}(\tau_t^\star)}
 =\prod_{h=1}^H
 \frac{\pi_h(a_{t,h}^\star\mid s_{t,h}^\star)}
 {\pi_h^\star(a_{t,h}^\star\mid s_{t,h}^\star)}.
\end{equation}
Thus action feedback suffices to evaluate
\cref{eq:offline-token-alg} without knowing the dynamics.

Under \cref{ass:ratio-cap}, with probability at least $1-\delta$,
\begin{align*}
 L(\widehat\pi)
 &\lesssim B_\infty\Hell^2(P^{\widehat\pi},P^{\pi^\star})
 \tag{\cref{lem:ratio-conversion}, $B_\infty>1$}\\
 &\lesssim B_\infty\left[
   \min_{\pi\in\Pi}\Hell^2(P^\pi,P^{\pi^\star})
   +\frac{\log(|\Pi|/\delta)}T\right]
 \tag{\citealp[Proposition~C.1]{rohatgi2025}}\\
 &\le B_\infty\left[L^\star+\frac{\log(|\Pi|/\delta)}T\right].
 \tag{\cref{eq:hellinger-kl}}
\end{align*}
\end{proof}


%% file: appendix_auxiliary.tex
\section{Auxiliary lemmas}\label{app:preliminaries}
\input{appendix_second_moment}

\subsection{KL and Hellinger inequalities}\label{app:divergence-tools}
All distributions below are on a common discrete space. We write
$P\ll Q$ when $Q(z)=0$ implies $P(z)=0$. Ratios $P(z)/Q(z)$
are evaluated where $Q(z)>0$, and equal zero when $P(z)=0$.

\begin{lemma}[KL variational inequality
{\citep[Theorem~4.6]{polyanskiy2024}}]
\label{lem:kl-variational}
Let $R,S$ be probability distributions with $\KL(R\|S)<\infty$.
For every real-valued function $g$ satisfying
$0<\E_S\left[e^g\right]<\infty$,
\[
 \KL(R\|S)
 \ge\E_R\left[g\right]-\log\E_S\left[e^g\right].
\]
\end{lemma}

\begin{lemma}[Elementary Hellinger inequalities]\label{lem:hellinger}
For probability distributions $P,Q$,
\begin{equation}\label{eq:hellinger-kl}
 \Hell^2(P,Q)=\sum_z(\sqrt{P(z)}-\sqrt{Q(z)})^2
 \le\KL(P\|Q).
\end{equation}
For conditional probabilities $r(y\mid x)\ge0$ with
$\sum_y r(y\mid x)=1$, define
$\widetilde P(y)=\sum_x r(y\mid x)P(x)$ and
$\widetilde Q(y)=\sum_x r(y\mid x)Q(x)$. Then
\begin{equation}\label{eq:auxiliary-hellinger-contraction}
 \Hell(\widetilde P,\widetilde Q)\le\Hell(P,Q).
\end{equation}
For a function $F$ with $0\le F(z)\le M$,
\begin{equation}\label{eq:weighted-hell}
 \sqrt{\E_P\left[F\right]}
 \le\sqrt{\E_Q\left[F\right]}+\sqrt M\,\Hell(P,Q).
\end{equation}
For distributions $R,S,S'$ and a subset $D$ of their common space,
\begin{equation}\label{eq:overlap-triangle}
 \sqrt{\sum_{z\in D}\min\{R(z),S(z)\}}
 \le\sqrt{\sum_{z\in D}\min\{R(z),S'(z)\}}+\Hell(S,S').
\end{equation}
\end{lemma}
\begin{proof}
For \cref{eq:hellinger-kl}, only $\KL(P\|Q)<\infty$ needs proof.
The scalar inequality
$(\sqrt x-1)^2\le x\log x-x+1$ for $x\ge0$ gives
\begin{align*}
 \Hell^2(P,Q)
 &=\E_Q\left[\left(\sqrt{P/Q}-1\right)^2\right]
 \tag{$P\ll Q$}\\
 &\le\E_Q\left[(P/Q)\log(P/Q)-P/Q+1\right]
 \tag{Scalar inequality}\\
 &=\KL(P\|Q).
 \tag{$\sum_zP(z)=\sum_zQ(z)=1$}
\end{align*}
For \cref{eq:auxiliary-hellinger-contraction},
\begin{align*}
 \Hell^2(\widetilde P,\widetilde Q)
 &=2-2\sum_y\sqrt{
   \Bigl(\sum_xr(y\mid x)P(x)\Bigr)
   \Bigl(\sum_xr(y\mid x)Q(x)\Bigr)}
 \tag{Definition of $\widetilde P,\widetilde Q$}\\
 &\le2-2\sum_y\sum_xr(y\mid x)\sqrt{P(x)Q(x)}
 \tag{Cauchy--Schwarz}\\
 &=2-2\sum_x\sqrt{P(x)Q(x)}=\Hell^2(P,Q).
 \tag{$\sum_y r(y\mid x)=1$}
\end{align*}
For \cref{eq:weighted-hell},
\begin{align*}
 \sqrt{\E_P\left[F\right]}
 &=\|\sqrt F\sqrt P\|_2
 \tag{Definition of $\|\cdot\|_2$}\\
 &\le\|\sqrt F\sqrt Q\|_2
       +\|\sqrt F(\sqrt P-\sqrt Q)\|_2
 \tag{Triangle inequality}\\
 &\le\sqrt{\E_Q\left[F\right]}+\sqrt M\,\Hell(P,Q).
 \tag{$0\le F\le M$}
\end{align*}
For \cref{eq:overlap-triangle}, the minimum is taken pointwise, and
\begin{align*}
 \sqrt{\sum_{z\in D}\min\{R(z),S(z)\}}
 &=\|\1_D\min\{\sqrt R,\sqrt S\}\|_2
 \tag{Definition of $\|\cdot\|_2$}\\
 &\le\|\1_D\min\{\sqrt R,\sqrt{S'}\}\|_2
       +\|\1_D(\sqrt S-\sqrt{S'})\|_2
 \tag{Triangle inequality and $1$-Lipschitz minimum}\\
 &\le\sqrt{\sum_{z\in D}\min\{R(z),S'(z)\}}
       +\Hell(S,S').
 \tag{$\1_D\le1$}
\end{align*}
\nopagebreak[4]
\end{proof}

\begin{lemma}[Hellinger-to-KL conversion]\label{lem:ratio-conversion}
If $P\ll Q$ and $P(z)/Q(z)\le e^b$ for $b\ge0$, then
\[
 \Hell^2(P,Q)\le\KL(P\|Q)\lesssim(1+b)\Hell^2(P,Q),
 \qquad \KL(P\|Q)\le b.
\]
\end{lemma}
\begin{proof}
For $0\le x\le e^b$, the scalar comparison underlying
\citet[Theorem~9]{sason2016}, at Hellinger order $1/2$, gives
$x\log x-x+1\lesssim(1+b)(\sqrt x-1)^2$.
Hence
\begin{align*}
 \Hell^2(P,Q)
 &\le\KL(P\|Q)
 \tag{\cref{eq:hellinger-kl}}\\
 &=\E_Q\left[(P/Q)\log(P/Q)-P/Q+1\right]
 \tag{$\E_Q\left[P/Q\right]=1$}\\
 &\lesssim(1+b)\E_Q\left[\left(\sqrt{P/Q}-1\right)^2\right]
 \tag{Scalar comparison, $P/Q\le e^b$}\\
 &=(1+b)\Hell^2(P,Q).
 \tag{Definition of $\Hell^2$}
\end{align*}
Also $\KL(P\|Q)=\E_P\left[\log(P/Q)\right]\le b$.
\end{proof}

\subsection{Concentration and testing bounds}\label{app:concentration-tools}
Let $(\mathcal F_t)_{t=0}^T$ be the training-history filtration.
The history $\mathcal F_{t-1}$ includes all observations and
randomization before episode $t$, and
$\E_t\left[\cdot\right]=\E\left[\cdot\mid\mathcal F_{t-1}\right]$.
In particular, $\pi_t$ and its trajectory are both averaged over
unless $\pi_t$ is explicitly included in the conditioning.

\begin{lemma}[A fixed-time exponential inequality]\label{lem:fixed-time}
Let $X_t$ be $\mathcal F_t$-measurable with $|X_t|\le2$.
For fixed $T$, $0<\lambda\le1/2$, and $\epsilon\in(0,1)$, each of
\begin{align}
 \sum_{t=1}^T X_t
 &\le\sum_{t=1}^T\E_t\left[X_t\right]
 +\lambda\sum_{t=1}^T\E_t\left[X_t^2\right]
 +\frac{\log(1/\epsilon)}{\lambda},\label{eq:exp-upper}\\
 \sum_{t=1}^T\E_t\left[X_t\right]
 &\le\sum_{t=1}^T X_t
 +\lambda\sum_{t=1}^T\E_t\left[X_t^2\right]
 +\frac{\log(1/\epsilon)}{\lambda}\label{eq:exp-lower}
\end{align}
holds with probability at least $1-\epsilon$.
\end{lemma}
\begin{proof}
The exponential inequality of \citet[Lemma~D.1]{foster2024},
applied to $-\lambda X_t$, gives, with probability at least
$1-\epsilon$,
\begin{align*}
 \sum_{t=1}^T X_t
 &\le\frac1\lambda\sum_{t=1}^T
       \log\E_t\left[e^{\lambda X_t}\right]
       +\frac{\log(1/\epsilon)}\lambda
 \tag{\citealp[Lemma~D.1]{foster2024}}\\
 &\le\frac1\lambda\sum_{t=1}^T
       \log\!\left(1+\lambda\E_t\left[X_t\right]
       +\lambda^2\E_t\left[X_t^2\right]\right)
 \notag\\[-2pt]
 &\qquad+\frac{\log(1/\epsilon)}\lambda
 \tag{$e^x\le1+x+x^2$, $|x|\le1$}\\
 &\le\sum_{t=1}^T\E_t\left[X_t\right]
       +\lambda\sum_{t=1}^T\E_t\left[X_t^2\right]
       +\frac{\log(1/\epsilon)}\lambda.
 \tag{$\log(1+x)\le x$}
\end{align*}
This is \cref{eq:exp-upper}. Applying the same inequality to
$-X_t$ gives \cref{eq:exp-lower}.
\end{proof}

\begin{lemma}[Empirical minimization of bounded scores]
\label{lem:bounded-score-erm}
Let the vectors $(Z_t(\pi))_{\pi\in\Pi}$ be independent and
identically distributed across $t$, with $0\le Z_t(\pi)\le1$ and
$\mu(\pi)=\E\left[Z_t(\pi)\right]$. Dependence across candidates
within an episode is allowed. Every empirical minimizer
$\widehat\pi\in\arg\min_{\pi\in\Pi}\sum_{t=1}^T Z_t(\pi)$
satisfies, with probability at least $1-\delta$,
\[
 \mu(\widehat\pi)\lesssim
 \min_{\pi\in\Pi}\mu(\pi)+\frac{\log(|\Pi|/\delta)}T.
\]
\end{lemma}
\begin{proof}
For every $\pi$, independence across episodes and boundedness give
$\E_t\left[Z_t(\pi)\right]=\mu(\pi)$ and
$\E_t\left[Z_t(\pi)^2\right]\le\mu(\pi)$.
With $\lambda=1/2$ and $\epsilon=\delta/(2|\Pi|)$,
\cref{eq:exp-upper} and \cref{eq:exp-lower} hold simultaneously
for all $\pi\in\Pi$ with probability at least $1-\delta$.
On this event,
\begin{align}
 \left|\frac1T\sum_{t=1}^T Z_t(\pi)-\mu(\pi)\right|
 &\le\frac1{2T}\sum_{t=1}^T\E_t\left[Z_t(\pi)^2\right]
       +\frac{2\log(2|\Pi|/\delta)}T
 \tag{\cref{eq:exp-upper,eq:exp-lower}}\\
 &\le\frac12\mu(\pi)+\frac{2\log(2|\Pi|/\delta)}T.
 &&(\text{$Z_t(\pi)^2\le Z_t(\pi)$})
 \label{eq:bounded-score-deviation}
\end{align}
For $\pi_\mu\in\arg\min_{\pi\in\Pi}\mu(\pi)$,
\begin{align*}
 \frac12\mu(\widehat\pi)
 &\le\frac1T\sum_{t=1}^T Z_t(\widehat\pi)
       +\frac{2\log(2|\Pi|/\delta)}T
 \tag{\cref{eq:bounded-score-deviation}}\\
 &\le\frac1T\sum_{t=1}^T Z_t(\pi_\mu)
       +\frac{2\log(2|\Pi|/\delta)}T
 \tag{Definition of $\widehat\pi$}\\
 &\le\frac32\mu(\pi_\mu)+\frac{4\log(2|\Pi|/\delta)}T.
 \tag{\cref{eq:bounded-score-deviation}}
\end{align*}
For $|\Pi|\ge2$, $\log(2|\Pi|/\delta)\lesssim\log(|\Pi|/\delta)$.
For $|\Pi|=1$, $\mu(\widehat\pi)=\min_{\pi\in\Pi}\mu(\pi)$.
\end{proof}

\begin{lemma}[Adaptive equality tests]\label{lem:equality-tests}
Let $\Theta$ be uniform on $[k]$ and $n\ge0$ be an integer.
A possibly randomized learner obtains information about $\Theta$
only through at most $n$ tests. Each test selects $j\in[k]$
using previous answers and reveals $\1\{j=\Theta\}$.
Every final estimate $\widehat\Theta\in[k]$ satisfies
\[
 \Prob\{\widehat\Theta\ne\Theta\}\ge1-\frac{n+1}{k}.
\]
\end{lemma}
\begin{proof}
The claim is trivial for $n\ge k-1$.
For a deterministic learner, let $j_1,\ldots,j_q$, $q\le n$,
and $b$ be its tests and final estimate when every answer is zero.
Then
\begin{align*}
 \Theta\notin\{j_1,\ldots,j_q\}
 &\quad\Longrightarrow\quad\widehat\Theta=b,
 \tag{Every test returns zero}\\
 \{\widehat\Theta=\Theta\}
 &\subseteq\{\Theta\in\{j_1,\ldots,j_q,b\}\},\\
 \Prob\{\widehat\Theta=\Theta\}
 &\le\frac{|\{j_1,\ldots,j_q,b\}|}{k}
 \le\frac{q+1}{k}\le\frac{n+1}{k}.
 \tag{Uniform $\Theta$}
\end{align*}
For a randomized learner, the same bound holds conditional on
its internal random choices and hence also after averaging.
\end{proof}

%% file: appendix_second_moment.tex
\subsection{Second moments and clipping}\label[appendix]{app:second-moment}

Throughout this subsection, write
$K_\pi(\tau):=\sum_{h=1}^H k_h(s_h;\pi)$.

\begin{lemma}[Sufficient conditions for the second-moment bound]
\label{lem:second-moment-sufficient}
Each of the following conditions is sufficient for
\cref{ass:relative-second-moment}.
\begin{enumerate}[leftmargin=1.5em,itemsep=0.2em,topsep=0.2em]
\item \textbf{Pathwise accumulated KL bound.}
If $B\ge1$ and, for every $\pi\in\Pi$ and
$\tau\in\mathcal T_{\rm env}$,
$K_\pi(\tau)\le B$, then \cref{ass:relative-second-moment} holds.
\item \textbf{Trajectory likelihood-ratio bound.}
\cref{ass:ratio-cap} implies \cref{ass:relative-second-moment}
with $B=4(B_\infty+1)$.
\end{enumerate}
\end{lemma}

\begin{proof}[Proof of \cref{lem:second-moment-sufficient}]
For the first claim, nonnegativity and the pathwise bound give
$K_\pi^2\le B K_\pi$. Taking expectation under $P^\pi$ yields
\[
\E_{P^\pi}[K_\pi^2]
\le B\,\E_{P^\pi}[K_\pi]
= B L(\pi).
\]

For the second claim, fix $\pi\in\Pi$ and write
\[
P=P^\pi,\qquad Q=P^{\pi^\star},\qquad
b=B_\infty,\qquad K=K_\pi.
\]
Abbreviate $k_h=k_h(s_h;\pi)$.
For $h=1,\ldots,H$, let $\xi_h=(s_1,a_1,\ldots,s_h)$
denote the history immediately before action $a_h$.
Let $\mathcal F_{h-1}=\sigma(\xi_h)$
and let $\mathcal F_H=\sigma(\tau)$.
For $h=0,\ldots,H$, define the prefix likelihood ratio
and its logarithm by
\[
R_h=\prod_{j=1}^h
\frac{\pi_j(a_j\mid s_j)}{\pi_j^\star(a_j\mid s_j)},
\qquad
X_h=\log R_h,
\]
where logarithms are evaluated $P$-almost surely.
The common initial-state distribution gives $R_0=1$ and $X_0=0$.
Moreover,
\[
R_h=\E_Q[R_H\mid\mathcal F_h]\le e^b,
\qquad\text{so}\qquad X_h\le b.
\]
Let $Y_h=X_h-X_{h-1}$. Since the common transition
probabilities cancel,
\[
Y_h
=
\log\frac{\pi_h(a_h\mid s_h)}
              {\pi_h^\star(a_h\mid s_h)},
\qquad
\E_P[Y_h\mid\mathcal F_{h-1}]=k_h.
\]

All subsequent quantities are integrable.
Indeed, the likelihood-ratio bound implies $P\ll Q$,
\[
\E_P[X_h^2]
=
\E_Q[R_h(\log R_h)^2]
\le b^2+\frac{4}{e^2},
\]
and $0\le k_h\le b-X_{h-1}$.
Since $H$ is finite, $K$ is square-integrable.

By the tower property,
\begin{align}
\E_P\!\left[
\sum_{j=h+1}^H k_j
\,\middle|\,\mathcal F_{h-1}
\right]
&=
\E_P[X_H-X_{h-1}\mid\mathcal F_{h-1}]
-k_h \notag\\
&\le b-X_{h-1}-k_h.
\label{eq:remaining-kl-bound}
\end{align}
Expanding $K^2$ and applying
\cref{eq:remaining-kl-bound} therefore gives
\begin{align}
\E_P[K^2]
&=
\E_P\!\left[
\sum_{h=1}^H
\left(k_h^2+2k_h\sum_{j=h+1}^H k_j\right)
\right] \notag\\
&\le
2\E_P\!\left[
\sum_{h=1}^H k_h(b-X_{h-1})
\right]
-\E_P\!\left[\sum_{h=1}^H k_h^2\right] \notag\\
&\le
2\E_P\!\left[
\sum_{h=1}^H k_h(b-X_{h-1})
\right].
\label{eq:cumulative-kl-second-moment}
\end{align}

We next use the scalar inequality
\begin{equation}
u(\log u)^2
\le(c+2)(u\log u-u+1),
\qquad
0\le u\le e^c,\quad c\ge0.
\label{eq:scalar-log-second-moment}
\end{equation}
At $u=0$, the left-hand side is defined as zero.
For $0<u\le1$, the stronger bound with coefficient $2$
follows from $e^t\ge1+t+t^2/2$, with $t=-\log u$.
For $u\ge1$, the stronger bound with coefficient $\log u+2$
follows from
\[
(\log u+2)(u\log u-u+1)-u(\log u)^2
=
(u+1)\log u-2(u-1)\ge0.
\]
This proves \cref{eq:scalar-log-second-moment}.

Conditional on $\mathcal F_{h-1}$, the action likelihood
ratio is bounded above by $e^{b-X_{h-1}}$.
Applying \cref{eq:scalar-log-second-moment} to this ratio,
with $c=b-X_{h-1}$, and averaging under the teacher's
conditional action distribution yields
\begin{equation}
\E_P[Y_h^2\mid\mathcal F_{h-1}]
\le (b-X_{h-1}+2)k_h.
\label{eq:conditional-log-second-moment}
\end{equation}
Telescoping $X_h^2$ and using
\cref{eq:conditional-log-second-moment}, we obtain
\begin{align*}
\E_P[X_H^2]
&=
\E_P\!\left[
\sum_{h=1}^H
\left(
2X_{h-1}k_h
+\E_P[Y_h^2\mid\mathcal F_{h-1}]
\right)
\right]\\
&\le
\E_P\!\left[
\sum_{h=1}^H (b+X_{h-1}+2)k_h
\right]\\
&=
2(b+1)L(\pi)
-
\E_P\!\left[
\sum_{h=1}^H k_h(b-X_{h-1})
\right].
\end{align*}
Since $\E_P[X_H^2]\ge0$, combining this inequality
with \cref{eq:cumulative-kl-second-moment} proves
\[
\E_{P^\pi}[K_\pi^2]
\le 4(b+1)L(\pi)
=4(B_\infty+1)L(\pi).
\]
\end{proof}

\begin{lemma}[Clipping under a relative second-moment bound]
\label{lem:second-moment-clipping}
Under \cref{ass:relative-second-moment}, every $\pi\in\Pi$ satisfies
$L(\pi)\le B$ and
\[
\overline L(\pi):=\E_{P^\pi}[\min\{1,K_\pi\}]
\le L(\pi)\le(1+B)\overline L(\pi)\le2B\overline L(\pi).
\]
\end{lemma}
\begin{proof}
Fix $\pi$ and write $K=K_\pi$ and $f=\min\{1,K\}$.
Jensen's inequality gives
$L(\pi)^2\le\E_{P^\pi}[K^2]\le BL(\pi)$, hence $L(\pi)\le B$.
The clipping inequality is immediate if $L(\pi)=0$.
Otherwise, set $K/\sqrt f=K^2/f=0$ when $K=0$.
Cauchy--Schwarz and $K^2/f\le K+K^2$ give
\[
\begin{aligned}
L(\pi)^2
&=\left(\E_{P^\pi}\left[\sqrt f\,\frac K{\sqrt f}\right]\right)^2
\le\E_{P^\pi}[f]\,\E_{P^\pi}[K^2/f]\\
&\le\overline L(\pi)\bigl(L(\pi)+\E_{P^\pi}[K^2]\bigr)
\le(1+B)\overline L(\pi)L(\pi).
\end{aligned}
\]
Dividing by $L(\pi)>0$ proves the upper bound.
The lower bound follows from $f\le K$, and $1+B\le2B$ since $B\ge1$.
\end{proof}